\def\year{2021}\relax
\documentclass[letterpaper]{article} 
\usepackage{aaai21}  
\usepackage{times}  
\usepackage{helvet} 
\usepackage{courier}  
\usepackage[hyphens]{url}  
\usepackage{graphicx} 
\usepackage{natbib}  
\usepackage{caption} 
\long\def\COMMENT#1\ENDCOMMENT{\message{(Commented text...)}\par}

\usepackage[dvipsnames]{xcolor}
\usepackage{booktabs}
\usepackage{amsmath}
\usepackage{amssymb}
\usepackage{enumitem}
\usepackage{adjustbox}
\usepackage[labelformat=simple]{subcaption}
\usepackage{makecell}
\usepackage{multirow}

\usepackage[switch]{lineno} 
\usepackage{listings}
\lstdefinelanguage{clingo}{
	keywordstyle=[1]\usefont{OT1}{cmtt}{m}{n},%
	keywordstyle=[2]\textbf,%
	keywordstyle=[3]\usefont{OT1}{cmtt}{m}{n},
	alsoletter={\#,\&},%
	keywords=[1]{not,from,import,exists,else,return,while,
		break,and,or},%
	keywords=[2]{precondition,effect,member,executable_if,initially,causes,fluent,action,if},%
	keywords=[3]{ },%
	morecomment=[l]{\#\ },%
	morecomment=[l]{\%\ },%
	commentstyle={\color{darkgray}}%
}
\title{
Can Transformers Reason About Effects of Actions?
}
\author{
Pratyay Banerjee,
Chitta Baral,
Man Luo,
Arindam Mitra\textsuperscript{2},
Kuntal Pal,\\
Tran C. Son\textsuperscript{3},
Neeraj Varshney\\
{Arizona State University},
{Microsoft}\textsuperscript{2},
{New Mexico State University}\textsuperscript{3}\\
pbanerj6,chitta,mluo26,kkpal,
nvarshn2@asu.edu,\\
arindam.mitra@microsoft.com, 
stran@nmsu.edu}

\begin{document}


\maketitle

\begin{abstract}
A recent work has shown that transformers are able to ``reason'' with facts and rules in a limited setting where the rules are natural language expressions of conjunctions of conditions implying a conclusion. Since this suggests that transformers may be used for  reasoning with knowledge given in natural language, we do a rigorous evaluation of this with respect to a common form of knowledge and its corresponding reasoning -- the reasoning about effects of actions. Reasoning about action and change has been a top focus in the knowledge representation subfield of AI from the early days of AI  and more recently it has been a highlight aspect in common sense question answering. 
We consider four action domains (Blocks World, Logistics, Dock-Worker-Robots and a Generic Domain) in natural language and create QA datasets that involve reasoning about the effects of actions in these domains. We investigate the ability of transformers to (a) learn to reason in these domains and (b) transfer that learning from the generic domains to the other domains.
\end{abstract}

\section{Introduction} \label{intro}

\COMMENT 
Since the advent of AI there has been efforts to design knowledge representation languages and reasoning mechanisms based on that. Research in this has been always motivated by examples given in natural language. While in the early days researchers took liberty of manually translating the natural language examples to symbolic representation, some researchers in NLP have been trying to automatically convert natural language text to symbolic representations so that they can reason with that. This approach, referred to as semantic parsing, has not been that successful. In this paper, we ask the question, Can we do away with the symbolic representation and just represent complex knowledge in natural language and reason with that. There are two advantages of this approach: (a) No need for semantic parsing, which is still not very accurate. (b) No need to manually write or learn rules in a formal logic such as ASP.

{\color{red}
Questions we answer:
\begin{enumerate}
    \item Can Transformers learn Reasoning about actions?
    \item Can it generalize to different and novel worlds?
    \item Can we learn from a specific General Domain and transfer to other specific Action Domains?
    \item Can a knowledge description of Action Laws, increase Few-Shot performance?
    \item Can it generalize to different Action depths, where the number of states increase after each action ?
    \item Can it learn numeric descriptive reasoning about action domain worlds?
\end{enumerate}
}
{\color{blue}
Implications of our work suggests opportunities to be explored in action-domain specific question answering, human-robot interactions, and an alternate perspective to knowledge representation and reasoning.
}
\ENDCOMMENT

Giving the long pursued goal of  AI - starting with McCarthy's Advice Taker \cite{mccarthy1959programs} - of having systems that can reason with explicitly given general knowledge as a motivation, the recent work \cite{clark2020transformers} studies the ability of transformers to ``reason’’ with facts and rules given as natural language sentences. They show that in their limited setting where rules are natural language expressions of conjunctive implications of the form ``[ $\wedge$ condition ]$^*$ $\rightarrow$ conclusion'' transformers can ``\textit{reason}’’ with such rules (and facts) given in natural language and answer yes/no questions with 99\% correctness. They also show that their model generalizes to test data that requires a much longer chain of reasoning with an accuracy of 95+\%. Intrigued by their result, our aim in this paper is to further study this approach with respect to reasoning with more general forms of knowledge.

Reasoning about action and change (RAC) is one of the key topic in knowledge representation and reasoning from the early days of AI. The ``\textit{going to the airport}'' example in \cite{mccarthy1959programs} involved reasoning about the effect of the actions ``\textit{walk to the car}'' and ``\textit{drive the car to the airport}''. Subsequently McCarthy and others developed the Situation calculus as a tool to reason about actions, with the first paper on this in 1963 \cite{mccarthy1963situations} and the seminal paper on it in 1969 \cite{mccarthy1969some} where the ``\textit{frame problem}'' was introduced. Since then there have been many specialized collections (workshop proceedings, journal special issues) on the topics of ``\textit{frame problem}'' and ``\textit{reasoning about actions and change}''; and also several books \cite{shanahan1997solving,reiter2001knowledge} on the topic. While ``\textit{reasoning about actions and change}'' continues to be an active area of research in the knowledge representation and reasoning community, recently it has been a highlight aspect in common sense question answering. For example, the ATOMIC knowledge base \cite{sap2019atomic} is about actions and their attributes such as (conditional) effects, executability conditions, triggering or preceding conditions, and motivations; and it is used as a key source of commonsense knowledge in the NLQA system \cite{bosselut2019comet}. 

Thus, in this paper, in our quest to extend the study in \cite{clark2020transformers} to more general forms of knowledge, we explore how well transformers can emulate ``\textit{reasoning}'' about effects of actions. We explore three example domains, Blocks World, Logistics and DWR (Dock-Worker-Robot) \cite{ghallabnt04} which we collectively refer to as the BLD domains, and a generic domain. The Blocks World domain is about a world where blocks are on a table or on top of another block and can be made to towers on the table with the actions being moving a block from its current position to another position. The Logistics domain is about
loading/unloading packages from trucks and airplanes and 
their movements from one position to another. The DWR domain is about a harbor that has locations, cranes, containers, pallets, piles and robots; where pallets can be made to piles, loaded in containers and robots, and moved by cranes; and robots can move from one location to another. The generic domain expresses effects of abstract actions on world properties. We create synthetic worlds of various complexity (based on the number of objects or actions) in these domains and create QA examples with respect to them that we divide into training and test sets. The QA examples includes items with yes/no answers as well as items with somewhat open-ended answers along with numerical values. To create the synthetic examples we use Answer Set Programming \cite{gelfond1988stable}, a declarative language that can express the frame problem in a natural way, and that has many solvers. In our synthetic examples, we focus on the simpler reasoning aspects such as starting with a completely known initial situation whether an action is executable in a particular situation, whether particular properties (referred to as``\textit{fluents}'') are true in a particular situation (reached after the execution of a sequence of actions), and count of such properties. Following is an example of QA items created by our synthetic approach:

\vspace{1pt}
\setlength{\fboxsep}{1pt}
\noindent
\resizebox{\linewidth}{!}{
    \fbox{
    \footnotesize
    \parbox[t]{\linewidth}
        {
            \textbf{Initial State}:  These are locations: fishery, airfield.  Robots are: robot-3, r10.  Crane are: crane-9, crane-7.  There are piles: pile-9, pile-12.  These are containers: seashell, moccasin.  fishery is adjacent to airfield.  fishery has the pile pile-9. pile-12 is at airfield.  crane-9 is located at fishery. airfield has crane-7.  robot-3 presents at airfield.  fishery has r10.    robot-3 is unloaded.  r10 can hold a container.  crane-9 can hold a container.  crane-7 can hold a container.  These are stacked in order top to bottom : seashell, moccasin.  pile-12 has seashell at the top.\\ 
            \textbf{Rules}: A robot can be at one location at a time. 
            A crane can move a container from the top of a pile to an empty robot or to the top of another pile at the same location.
            A container can be stacked in some pile on top of the pallet or some other container, loaded on a robot, or held by a crane.
            A pile is a fixed area attached to a single location.
            A crane belongs to a single location; it can manipulate containers within that location, between piles and robots.
            Each robot can carry one container at a time.
            Robots can move to free adjacent location.
            A crane is empty if it is not holding container.
            A robot is unloaded if it is not loaded with container.\\
            \textbf{Action}: Crane-7 picks up seashell and seashell is loaded on r10. \\
            \textbf{Verify}~~Q:) \textit{Is r10 loaded with moccasin?}\;\;Ans: \textbf{No}.\\
            \textbf{Counting}~~Q:) \textit{how many robots are unloaded?} Ans: \textbf{1}.\\
            \textbf{Others}~~Q:) \textit{which container is on top of pile-12?} Ans: \textbf{moccasin}.
        }
    }
}

We pursue various experiments on our synthetic dataset to answer questions and analyze how well a transformer fine tuned with the dataset of \cite{clark2020transformers} can reason about effect of actions in the BLD domains, how well a transformer fine tuned with our training sets does on our test sets, how well it generalizes when the domain is enhanced with more objects, and how well a transformer fine tuned with the generic domain transfers to the specific BLD  domains. In our study we consider both explicit specification of rules that express effect of actions as well as the case when these effects are not explicitly given and learned. 


Our contributions are summarized below:
\begin{enumerate}[noitemsep]
    \item We provide a framework to study the extent of neural NLQA model's ability to reason about effect of actions. The framework consists of procedurally generated question answering dataset with three types of questions, on four action domain worlds out of which, two are real-world  domains. We also provide a hand authored test set.
    \item We show RuleTakers \shortcite{clark2020transformers}, trained on conjunctive implication rules is unable to generalize to such questions about effect of actions (48-58\%).
    \item We perform extensive experiments to study the out-of-domain generalization abilities of current state-of-the-art transformer-encoder based QA model.
    \item We observe transformer-encoder based QA models can perform reasoning about effect of actions, and do generalize to some extent to out-of-domain worlds (68-90\%), but there is still scope of significant improvements.
    \item We also investigate the ability to learn from a generic domain of actions and fluents, and  observe models can somewhat generalize to real world domains (57-83\%).
\end{enumerate}

\begin{figure}[t]
    \centering
    \includegraphics[height=1.5in,width=\linewidth]{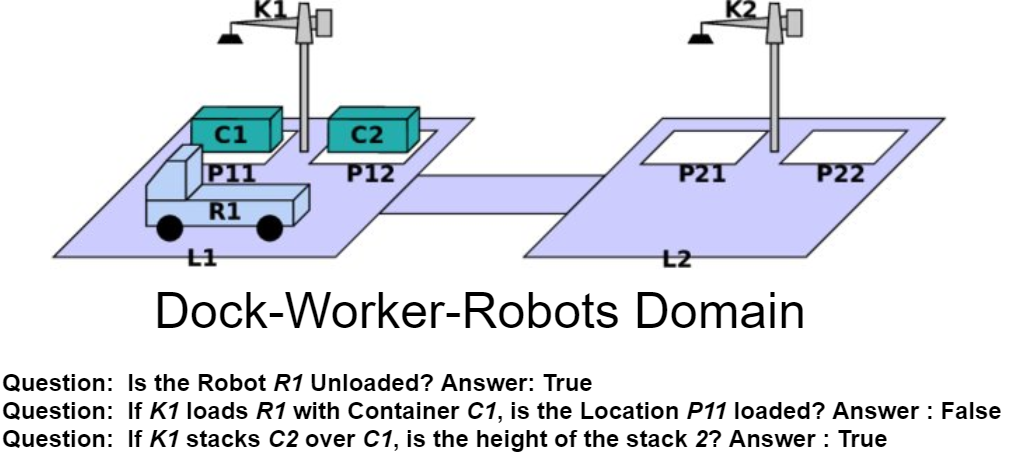}
    \caption{Reasoning about Actions using Question Answering. The world is described in procedurally generated text.}
    \label{fig:te}
\end{figure}

\COMMENT

\begin{enumerate}

\item  We show that Roberta \cite{liu2019roberta} fine tuned with the dataset in \cite{clark2020transformers} perform randomly on True/False questions on effects of actions in the Blocks World and Logistics domain and slightly better (58\% accuracy) in the DWR domain.

\item  If Roberta is fine tuned with our Blocks World without the action effect rules dataset then it performs nearly perfect (99.5+\%) in domains with the same number of towers and the performance gradually degrades as the number of towers in the test set are increased. We see similar results in the logistics domain. 

\item  If Roberta is fine tuned with our Blocks World with the action effect rules dataset then also it performs nearly perfect (99.5+\%) in domains with the same number of towers and the performance gradually degrades as the number of towers in the test set are increased. 

(a) When there is a bigger difference in the number of blocks between the training set and test set, the generalization in the Blocks world domain  is better when the action effect rules are specified. 

(b) We also observe that presence of the action effect rules leads to greater accuracy with smaller size training sets. 

\item  Compared to Novel World Generalization, Action depth generalization is observed to be harder. With action effect axioms, we observe slightly better performance. The task of open-ended Other questions, are comparatively harder than Binary and Counting questions.  The generalization accuracy drops faster without effect axioms. 

\item If Roberta is finetuned with our DWR without the action effect rules dataset then it performs around 80\% in low level complexity (with 2-4 objects of each class)  and  around 75\% in high level complexity (with 2-4 objects of each class). Presence of action effect rules do not make any consistent difference in the DWR domain.  

\item With increasing number of DWR data, Roberta performs better  
In the DWR domain, with the same level of complexity in the training and testing set, as the complexity increases the accuracy decreases.  However, if the training set has a higher complexity than the test set then the accuracy is higher than when the test set complexity is same. 

\item In DWR Roberta with fine tuning does not do well when the questions are open-ended. 

\item In the logistic domain with explicit knowledge the model is not able to generalize for higher complexity domains than what it has been trained on and we see a substantial drop in performance as the complexity increases.  In low complexity domains the model performance increases with knowledge, but as the complexity increases the performance drops with incorporation of knowledge.

\item In the generic domain We observe that the model performs significantly better than the random baseline(~80\%). A model trained on lower complexity domains (lower number of fluents and actions) generalizes well to higher complexity domains (higher number of fluents and actions) but its performance is slightly worse than the performance of its counterpart model trained on higher complexity domains. 
The model is able to generalize to some extent even with a small number of examples (better than random). The model attains its near optimal performance with 10000 data samples as the accuracy doesn’t increase significantly on training with larger datasets. 

\item Finally, when we convert all the actions domains (Blocks world, Logistics, DWR) to a generic format and evaluate its performance on a model trained on generic domain, we observe that the model trained on generic domain performs well on all the converted action domains. 

\end{enumerate}

\ENDCOMMENT

\section{BLDG: Domains}

 In this section, we describe the four action domains, the \emph{blocks world}, \emph{logistics}, \emph{dock-workers robots}, and \emph{generic} domain. The first two domains are well-known benchmarks in the classical automated planning competitions\footnote{\url{https://www.icaps-conference.org/competitions/}}. The third domain is from the text book by \citet{ghallabnt04}. We create the fourth domain to allow using a set-theoretic representation that enables simple generation of instances.

\subsection{Blocks World}
\label{blockworld}

A blocks world domain consists of a set of named cubes of the same size. A block can be on the table or on top of another block. A block is said to be clear if no other block is on top of it. A clear block can be moved to the table or on top of another clear block. No two blocks can be on the same block at a same time.
An \emph{instance} in this domain describes the names and location of the blocks (a.k.a. the \emph{initial state}).
Listing~\ref{lst:blocks} describes the effects of actions and their executability conditions of the actions in this domain. In this paper, we use the action language $\mathcal{A}$ proposed by \citet{gelfond1993representing} to represent the action domains.  


\begin{lstlisting}[language=clingo,caption=A Blocks World Domain, label=lst:blocks, 
mathescape=true,xleftmargin=.04\textwidth, breaklines=true]
block(a).   block(b).   ... 
fluent: on(X,Y), ontable(X), clear(X)  
action: move(X,Y), move(X,table)       
move(X,Y)  executable_if clear(X), clear(Y) 
move(X,ta) executable_if clear(X)
move(X,Y)  causes on(X,Y),$\neg$clear(Y)
move(X,Y)  causes clear(Z) if on(X,Z)
move(X,Y)  causes $\neg$on(X,Z) if on(X,Z)
move(X,Y)  causes $\neg$ontable(X) if ontable(X)
move(X,ta) causes ontable(X)
move(X,ta) causes clear(Z) if on(X,Z)
move(X,ta) causes $\neg$on(X,Z) if on(X,Z)
initially: ontable(b), on(a,b), ...
\end{lstlisting}

In the above, $X$, $Y$, and $Z$ stand for pairwise different blocks and $ta$ stands for table. The first line specificies the objects (blocks) in the instance, the second line defines the fluents, and the third line defines the actions of the domain. Lines 4 and 5 state the executability of actions. Lines 
6--12 encode the effects of the actions, e.g., Line 6 states two unconditional effects of the action of moving the block $X$ on top of the block $Y$: it will cause $X$ to be on $Y$ and $Y$ no longer clear; Line 7 specifies a conditional effect of the same action: it will clear the block $Z$ if $X$ is currently on $Z$, etc. The last line specifies the initial state of the world, the block $b$ is on the table and $a$ is on $b$, etc.

\subsection{Logistics}
\label{logistics}

A logistics domain consists of trucks, airplanes, packages at different locations, and different cities. Each location is associated to a city. A location can also be an airport. A package can be loaded to a truck (or an airplane) if the former is at the same location as the latter. Once a package is loaded into a truck (or an airplane), it is inside the truck (the airplane). If a package is inside a truck (an airplane), it can be unloaded from the truck (the airplane). Once unloaded, it is at the same location of the truck (the airplane). A truck can move from one location to another location within a city. An airplane can fly from one airport to another airport. It is assumed that a truck (an airplane) can hold unlimited number of packages. The initial state specifies the location of the trucks, airplanes, and packages.



% ...

\subsection{Dock-Workers Robots}
\label{dwr}

The dock-workers robots (DWR) domain could be viewed as a combination of the blocks world and the 
logistics domains without airplanes and airports and trucks are replaced by robots. The domain, taken inspiration from the arrangement of a habor, contains locations, cranes, containers, pallets, piles, and robots. Static information is provided describing the connectivity between locations, the attachment of pallets to locations, and the locations of cranes (that do not move). Robots can move from one location to the adjacenct one if it is not occupied by another robot. Containers are stacked into piles resembling a tower of blocks. A crane can take the container on top of a pile at the same location and holds it. It can also unload the container loaded on a robot at the same location. A crane can load the container it is holding on top of an empty robot or putdown the container on top of a pile at the same location. The initial state specifies the layout of the habor, i.e., the locations of all the objects and other properties (e.g., whether a robot is loaded with a container). 
Unlike other domains, the DWR domain contains state constraints (a.k.a. static causal laws) which express relationships between fluents\footnote{The usefulness of state constraints in planning has been discussed in \cite{thiebauxhn03,sontgm05a}}.



\subsection{Generic Domain}
\label{generic}

In this domain, an action is described by a set of facts of the forms in Listing~\ref{lst:action}. The first 
line defines the action $name$. The second line specifies that the executability condition 
of $name$ is a set called $pre(name)$. The third line declares that the elements of 
$pre(name)$ are the literals $f1,\ldots,fn$. The fourth line declares that the effect, 
coded $(name,1)$ is $f$. The sixth line specifies the conditions under which the effect $(name,1)$ 
is realized, which is the set of literals named $c(name,1)$ and contains $g1,\ldots,gn$. 
Similarly,  a state constraint is represented by a consequence (e.g., the fluent $occupied(L)$)  and a set associated with it (e.g., the set $\{at(R,L)\}$) via a keyword named $static\_law$. We omit the description of static causal laws here for brevity.  

\begin{lstlisting}[language=clingo,caption=Action Representation, label=lst:action, 
mathescape=true,xleftmargin=.03\textwidth, breaklines=true]
action(name) 
precondition(name, pre(name), _)
member(pre(name),f1) ... member(pre(name),fk).
effect(name, e(name,1), f) ....
member(c(name,1),g1) ... member(c(name,1),gn)
\end{lstlisting}

The advantage of this representation is that it enables the generation of random domains with a few parameters such as the number of fluents, the number of actions, the maximal numbers of effects of actions, the number of static causal laws, etc. 
Observe that this representation can be used to represent the three domains Blocks World, Logistics, and DWR. However, the size of the corresponding representation will increase dramatically, in terms of number of facts. For example, for the Block Worlds domains, with 7 blocks and three towers of 3, 1, 3 blocks, the generic 
representation has about 1500 facts. Observe that the number of actions and the number of fluents do not change in both representations.

\subsection{RAC in Logic Programming}
\label{rac-asp}

Given an action domain $D$, represented as a collection of statements about effects of actions and relationships between fluents (Subsections~\ref{blockworld}--\ref{generic}), an initial state $I$, stating the truth value of the fluents in the domain, and an integer $n$ denoting the length of the trajectories that we are interested in, a logic program, $P(D,I,n)$, can be automatically generated (see, e.g., \citet{GelfondL98}) that contains rules for
\begin{itemize}[noitemsep]
    \item  solving the three fundamental problems in reasoning about effects of actions: the \emph{qualification} problem (when can an action be executed?), the \emph{ramification} problem (how to deal with indirect effects of actions, represented by static laws, e.g., in DWR domain), and the \emph{frame} problem (how to deal with inertial?); 
    \item generating action occurrences: 
\end{itemize}
    \[
     {1\;\{occ(A,T):action(A)\}\;1:-time(T),T<n.}
    \]
The answer sets of $P(D,I,n)$ correspond one-to-one to possible evolutions of the world after $n$ actions (a.k.a. the trajectories of length $n$). Each answer set contains information about the executability of actions at each time step (e.g., whether an action of moving a block $a$ to be on top of the block $b$ can be executed after the execution of moving $b$ on top of $a$), the truth value of a fluent at a time step $k$ after the action sequence leading to the step $k$ has been executed (e.g., whether the block $a$ is on the table after moving on top of the block $b$, then moving the block $c$ on top of the block $d$). This, together with the fact that efficient and scalable ASP solver exists \textit{clingo} \cite{GebserKNS07}, allows for the automatic generation of QA examples for RAC by randomly generating the initial state $I$ and using $P(D,I,n)$ to generate different worlds. For example, with 20 blocks stacked up to create four 
towers and the length of 5 actions, generating an answer set (a world) with the corresponding questions can be done in less than 20 second.   
\section{Synthetic Data Generation}



In order to study transformer's capability of reasoning about effects of actions, we generate data for four action domains described in Subsections~\ref{blockworld}--\ref{generic} with three types of questions and different complexities. Each example is a triple (\textit{paragraph}, \textit{question}, \textit{answer}), where \textit{paragraph} provides the initial state, the knowledge of a domain, and a valid action sequence. We generate three types of questions based on the effects of actions to test the model's different reasoning abilities: \textit{verify}, \textit{counting}, and \textit{others}. A {\bf Verify question} is to infer if a statement can be entailed by the paragraph or not (e.g., ``Is it possible to move block A on top of block B?''), and answer can be either true or false. 
A {\bf Counting question} is to count the number of objects satisfying an attribute or a relation (e.g., ``How many blocks are on the table?''), and the answer is numeric value. 
An {\bf Others question} is to reason the status of objects (e.g., ``What is the position of block A?''), and the answer is an English word.

\subsubsection{Overview}
%
%
We randomly generated the initial states and use the method described in Subsection~\ref{rac-asp} to generate answer sets representing trajectories of length up to 5 for each action domain described in Subsections~\ref{blockworld}--\ref{generic}. 
We use the answer sets to generate questions of the three aforementioned types. To investigate the out-of-distribution (OOD) generalizability of a transformer, for each domain we define five levels of complexity in terms of \textit{novel world} and \textit{action depth}. Novel World complexity is defined by the number of towers for block world domain; the number of airplanes, trucks, and packages for logistic domain; the number of locations, cranes, robots and containers for dock-worker robots domain; the number of fluents, actions, and the complexity of the actions (the maximal number of fluents in the precondition, effects, conditions on each effects) for generic domain. Action Depth complexity is defined by the length of the interested trajectories. 


\subsubsection{World Generation}
World description includes the initial state and domain knowledge, where the former is different for each example, and the latter is the same for each example in each domain. The initial state includes two types of facts, attribute $object_i(n)$ (e.g., location(market)) and relation $rel_j(o_1, o_2)$ ( e.g., adjacent(market, park)).

The attribute predicates define objects in a domain or describe the status of objects and the relation predicates define the relation between objects. For four domains, the attributes and relation predicates are different:

\vspace{1pt}
\noindent
\resizebox{\linewidth}{!}{
\small
    \fbox{
        \parbox{\linewidth}{
            \textbf{Blocks world}: block/1, ontable/1, clear/1, on/2 ...\\
            \textbf{Logistic}: airplane/1, city/1, truck/1, location/1, inCity/2, ...\\
            \textbf{DWR}: location/1, crane/1, robot/1, pile/1, attached/2 ...\\
            \textbf{Generic}: action/1, fluent/1, effect/3, precondition/3, member/2 ...
        }
    }
}

where \textit{predicate/n} represents \textit{predicate} is an attribute or relation with $n$ parameters.  
The domain knowledge includes the effects and constraints of actions mentioned in previous sections. 





\subsubsection {Question Generation} 
For each domain, we generate three types of questions based on the effects of actions. \textit{Verify} questions are essentially binary questions. We generate ``true" questions by entailed effects, (e.g., by $empty(crane_1)$, a true question can be ``Is $crane_1$ empty?''). A ``false" question can be generated by negating the attribute of objects or replacing an object of a relation with others, (e.g., by $at(robot_1, market)$ in DWR, a false question can be ``Is $robot_1$ at airport?"). We balance the number of ``true'' and ``false'' questions to avoid label bias. To generate \textit{Counting} questions, we define a set of conditions for four domains, and count how many objects satisfy each condition, e.g. (given a condition in blocks world: executable actions at time 1, and a list of facts $executable(A_1, 1), executable(A_3, 1)$, a counting question with answer 2 can be ``How many actions are executable at time 1?''). 
To generate \textit{Others} questions, we select relation predicates, and mask one object in the question, e.g. (given a relation in logistic: $in(object_1, airplane_1)$, a question with answer $airplane_1$ can be ``Where is the $object_1$?''). We have a set of templates with different linguistic variations for each type of questions from which we randomly sample. More examples are  present in the supplemental material.


Since we procedurally generate the question-answer, we are able to generate a large number of train and test samples. We divide the train and test samples by first generating a set of unique  worlds with a pre-determined initial state configuration, to avoid train-test world overlaps. Further splits are made to test the different complexity worlds. 

\begin{figure*}[t]
\centering

    \begin{subfigure}{28mm}
        \includegraphics[width=\linewidth,height=25mm]{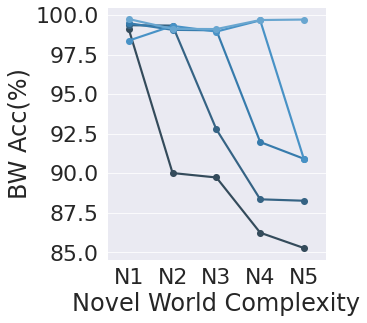}
    \end{subfigure}
    \begin{subfigure}{28mm}
         \includegraphics[width=\linewidth,height=25mm]{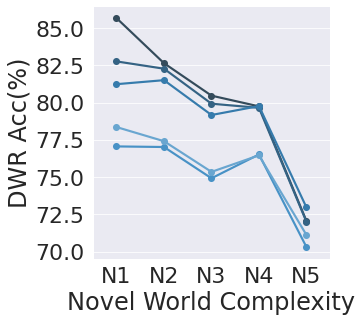}
    \end{subfigure}
    \begin{subfigure}{28mm}
         \includegraphics[width=\linewidth,height=25mm]{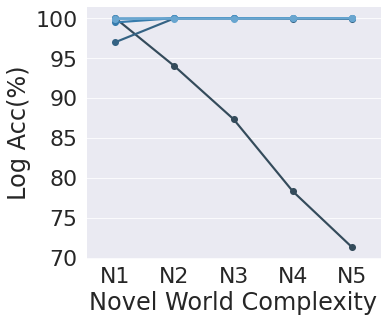}
    \end{subfigure}
    \begin{subfigure}{32mm}
         \includegraphics[width=\linewidth,height=25mm]{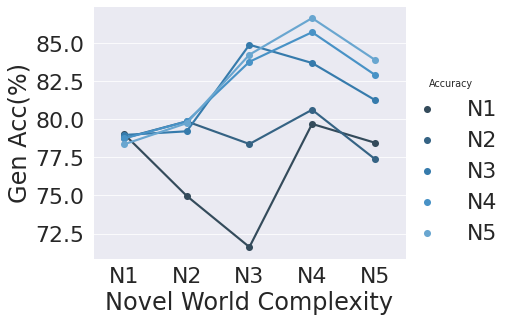}
    \end{subfigure}
    
    
    \begin{subfigure}{28mm}
        \includegraphics[width=\linewidth,height=25mm]{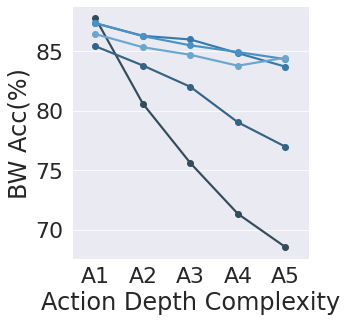}
    \end{subfigure}
    \begin{subfigure}{28mm}
         \includegraphics[width=\linewidth,height=25mm]{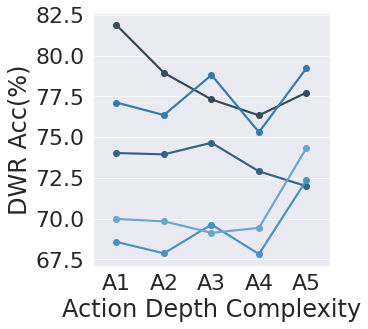}
    \end{subfigure}
    \begin{subfigure}{28mm}
         \includegraphics[width=\linewidth,height=25mm]{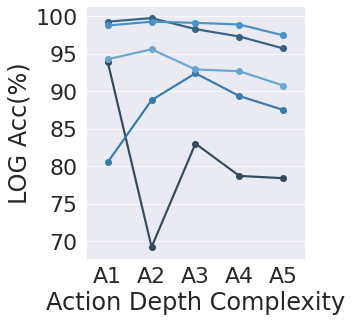}
    \end{subfigure}
    \begin{subfigure}{32mm}
         \includegraphics[width=\linewidth,height=25mm]{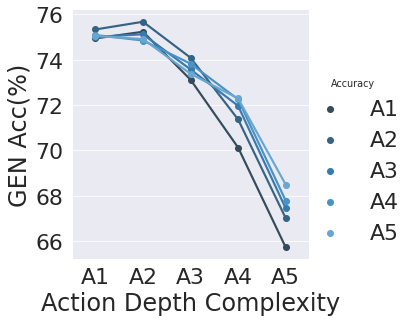}
    \end{subfigure}
    \caption{Accuracy trend when trained on simpler worlds  and tested on OOD complex worlds \textbf{with} action-effect axioms.}
    \label{fig:knowtrend}    
\end{figure*}

\begin{figure*}[t]
\centering
    \begin{subfigure}{28mm}
        \includegraphics[width=\linewidth,height=28mm]{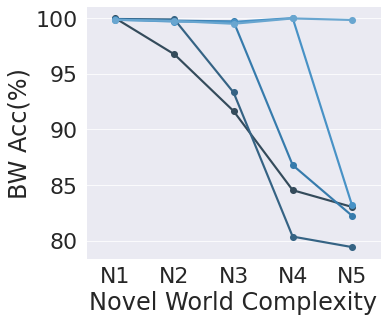}
    \end{subfigure}
    \begin{subfigure}{28mm}
         \includegraphics[width=\linewidth,height=28mm]{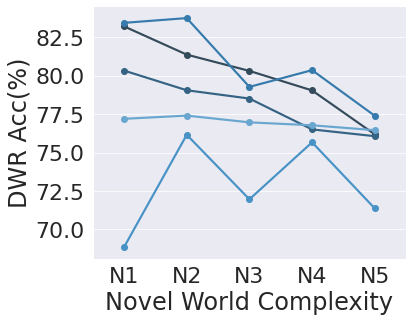}
    \end{subfigure}
    \begin{subfigure}{30mm}
         \includegraphics[width=\linewidth,height=28mm]{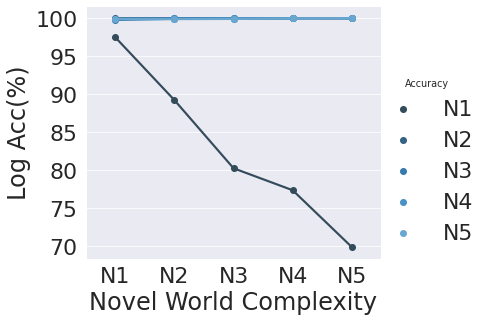}
    \end{subfigure}
    \begin{subfigure}{28mm}
        \includegraphics[width=\linewidth,height=28mm]{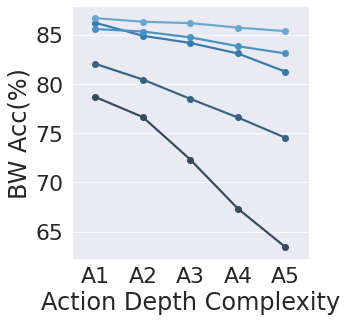}
    \end{subfigure}
    \begin{subfigure}{28mm}
         \includegraphics[width=\linewidth,height=28mm]{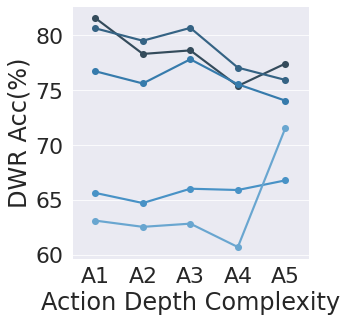}
    \end{subfigure}
    \begin{subfigure}{28mm}
         \includegraphics[width=\linewidth,height=28mm]{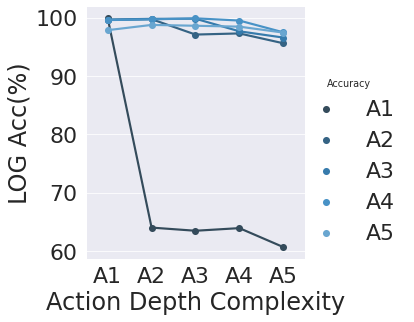}
    \end{subfigure}
    \caption{Accuracy trend when trained on simpler worlds  and tested on OOD complex worlds \textbf{without} action-effect axioms.}
    \label{fig:noknowtrend}    
\end{figure*}

\begin{figure*}[t]
    \begin{subfigure}{.21\textwidth}
        \includegraphics[width=\linewidth,height=27mm]{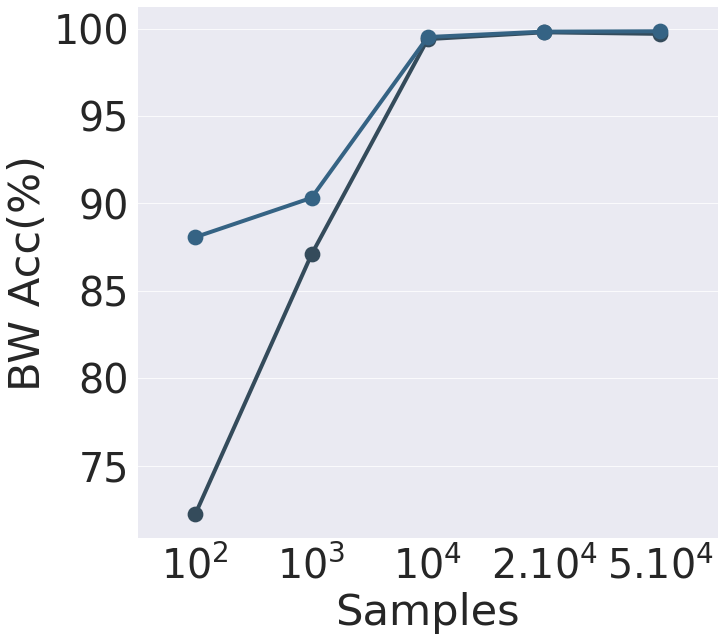}
    \end{subfigure}
    \begin{subfigure}{.21\textwidth}
         \includegraphics[width=\linewidth,height=27mm]{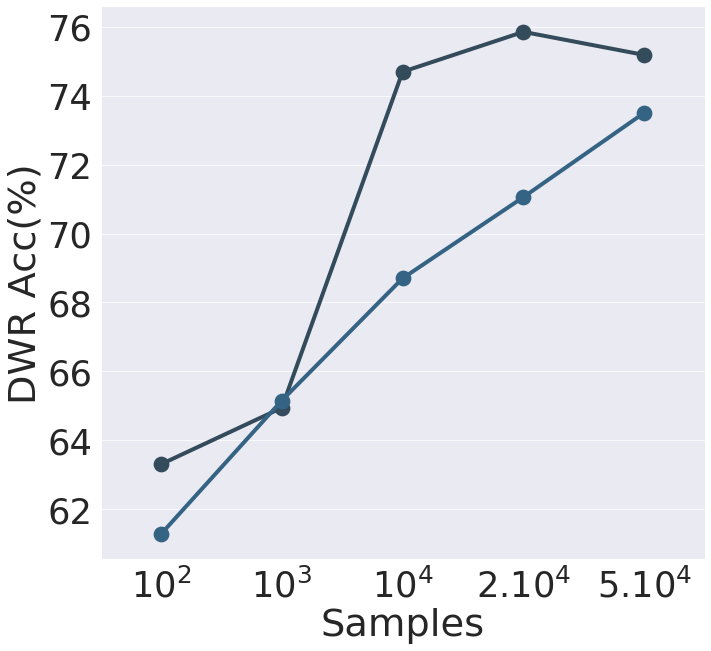}
    \end{subfigure}
    \begin{subfigure}{.30\textwidth}
         \includegraphics[width=\linewidth,height=27mm]{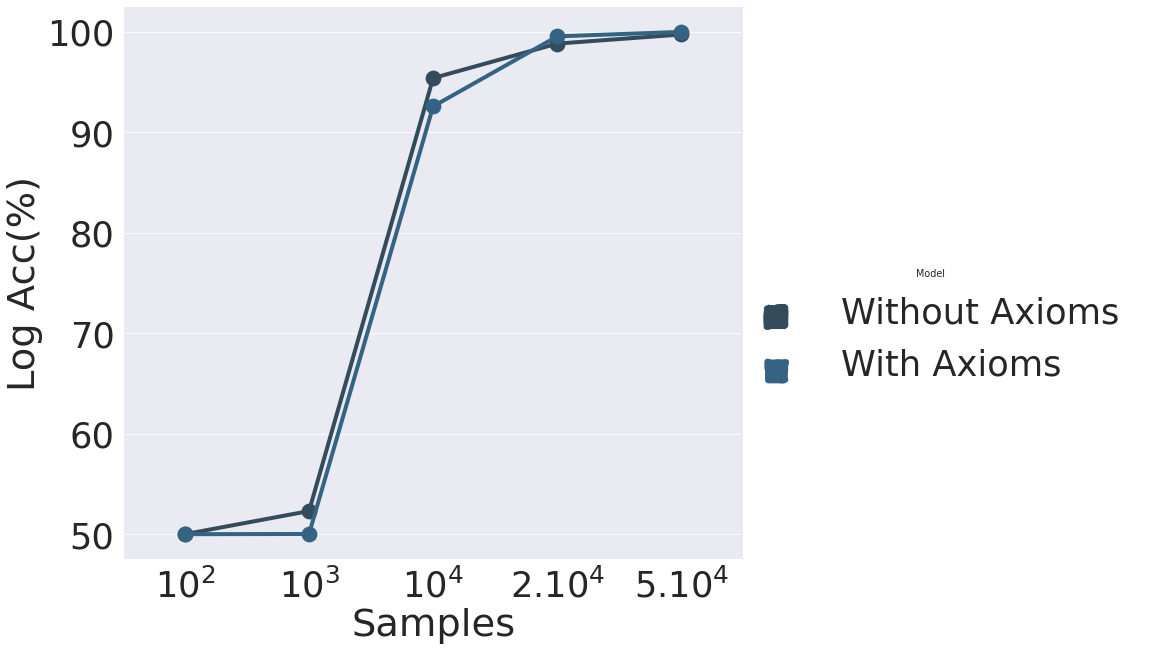}
    \end{subfigure}
    \begin{subfigure}{.26\textwidth}
         \includegraphics[width=\linewidth,height=27mm]{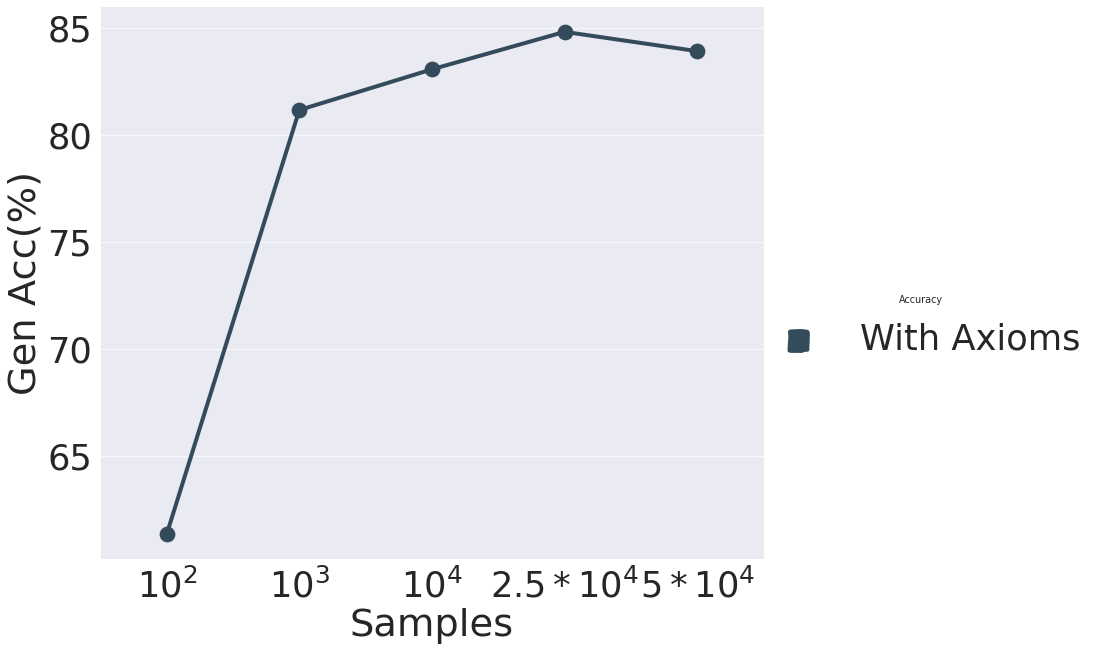}
    \end{subfigure}
    \caption{Learning curve trends for different domains on N5 data. For BLD we compare with and  without action-effect axioms.}
    \label{fig:learningcurve}    
\end{figure*}

\section{Experiments and Results} \label{experiments}
\subsection{Models}
\subsubsection{RuleTakers} We choose Roberta \cite{liu2019roberta} as the pretrained transformer-encoder based question answering model, as it has  been demonstrated by \citeauthor{clark2020transformers} to be a near perfect reasoner over natural language rules. We finetune the Roberta model following the recommended hyperparameters in \cite{clark2020transformers} on the associated reasoning dataset. We refer to this model as the RuleTaker. As the provided dataset has only True-False questions, we evaluate it on our Verify questions for all the domains. 

\subsubsection{Roberta with/without Action-Effect Rules} We also train the Roberta model on our generated dataset. We concatenate the initial state and sequence of actions (if any) alongwith the question and train two models, a True-False model similar to the RuleTakers on the binary classification task, and an Open-Ended QA model with a multi-class classification objective, where it learns to generate the answer (Yes/No, Number, Others) similar to the open-ended QA task of visual question answering \cite{antol2015vqa}.The answer vocabulary for the open-ended QA task is fixed for the train and test splits, and we ensure there are no unknown answers to avoid label shift. Questions are input as: $[\textit{CLS}] \textit{context} [\textit{SEP}] \textit{question} [\textit{SEP}]$, the [CLS] token representation is fed to a feedforward layer to project either to single logit for the binary classification task, or the answer vocabulary size for the open-ended QA task. The action-effect rules are concatentated with the initial state to create the context.  We use the standard cross-entropy loss to train. We evaluate the models using answer exact-match accuracy. 

\begin{table}[t]
\centering
\resizebox{\linewidth}{!}{
\begin{tabular}{@{}lcccc@{}}
\toprule
\textbf{Model} & \textbf{Blocks World $\uparrow$} & \textbf{Logistics $\uparrow$} & \textbf{DWR $\uparrow$} & \textbf{Generic $\uparrow$} \\ \midrule
RuleTakers & $51.0^{\dagger}$ & $48.0^{\dagger}$ & $58.0^{\dagger}$ & $53.01^{\dagger}$ \\
N1~+~Rules & \underline{85.3} & \underline{71.3} & 73.4 & 78.5 \\
N1~+~NoRules & 83.0 & 68.9 & \underline{76.2} & N/A \\
N3~+~Rules & 82.3 & \textbf{99.9} & 73.9 & \textbf{81.3 }\\
N3~+~NoRules & \textbf{90.9} & \textbf{99.9} & \textbf{77.4} & N/A \\ \midrule
Generic N1 & $63.4^{\dagger}$ & $57.2^{\dagger}$ & $78.8^{\dagger}$ & $78.4^{\dagger}$ \\
Generic N3 & $64.5^{\dagger}$ & $58.1^{\dagger}$ & $80.6^{\dagger}$ & $81.2^{\dagger}$ \\
Generic N5 & $70.5^{\dagger}$ & $59.6^{\dagger}$ & $83.5^{\dagger}$ & $83.9^{\dagger}$ \\ \bottomrule
\end{tabular}
}
\caption{Accuracy on T/F questions on OOD novel world (N5) test scenarios. Rules/NoRules indicates with and without action-effect axioms. $\dagger$ evaluated on N1 complexity. }
\label{tab:maintable}
\end{table}

\subsubsection{Training \& Testing}
All our experiments are conducted using
Roberta-large, which is finetuned on RACE dataset \cite{lai2017race} using the hyperparameters mentioned in \cite{liu2019roberta} as recommended by \citeauthor{clark2020transformers}. We use the same fixed hyperparameters (batch size, learning rate) and use 4 Nvidia V100 16GB gpus. 

To test the OOD zero-shot performance we defined the following world complexities, the novel worlds complexities (N1-N5) and action depth complexities (A1-A5). The novel worlds for Blocks world are defined using the number of towers (2,3,4,5,6). In DWR domain the locations, cranes, robots and containers with values for each in (2,3,4,5,6), and similarly for Logistics domain, on airplanes, trucks and packages with values for each in (2,3,4,5,6). In Generic domain the complexity is defined on number of  unique  actions and fluents with values in range (1-4,5-7,8-10,11-15,16-20). Action depth world complexity is defined the same throughout, with number of actions equal to (1,2,3,4,5). We train all our models with 40K balanced \textit{Verify} questions, 30K \textit{Counting} questions and 30K \textit{Others} questions. 

\begin{table*}[t]
\centering
\small
\begin{tabular}{@{}lccccccccccc@{}}
\toprule
 & \multicolumn{3}{c}{\textbf{Blocks World $\uparrow$}} & \multicolumn{3}{c}{\textbf{Logistics $\uparrow$}} & \multicolumn{3}{c}{\textbf{DWR  $\uparrow$}} & \multicolumn{2}{c}{\textbf{Generic $\uparrow$}} \\ 
\textbf{Model} & \textbf{T/F} & \textbf{Num} & \textbf{Other} & \textbf{T/F} & \textbf{Num} & \textbf{Other} & \textbf{T/F} & \textbf{Num} & \textbf{Other} & \textbf{T/F} & \textbf{Num} \\ \midrule
A1~+~Rules & \underline{75.13} & \underline{81.08} & \underline{47.18} & \underline{97.98} & \underline{55.20} & \underline{75.43} & \textbf{78.51} & \underline{80.30} & 74.12 & 87.67 & 36.47 \\
A1~+~NoRules & 72.40 & 76.31 & 38.61 & 75.29 & 33.85 & 68.12 & 78.38 & 78.73 & \textbf{74.88} & N/A & N/A \\
A3~+~Rules & \textbf{92.94} & \textbf{94.94} & \textbf{60.20} & 98.07 & 65.30 & 95.60 & 76.84 & \textbf{87.51} & 74.14 & \textbf{87.73} & \textbf{40.40} \\
A3~+~NoRules & 90.20 & 90.81 & 59.77 & \textbf{99.62} & \textbf{90.48} & \textbf{98.70} & 75.95 & 83.93 & 61.66 & N/A & N/A \\ \bottomrule
\end{tabular}
\caption{Accuracy on the three types of questions evaluated on OOD action depth A5 test scenarios.}
\label{tab:typeoodaction}
\end{table*}

\subsection{Results}

\subsubsection{Can RuleTakers reason about effect of actions?}
We evaluate the RuleTakers model on our generated \textit{Verify} questions with our action-effect axiom rules. The results are presented in Table \ref{tab:maintable}. We observe that it performs similar to near random in the Blocks World and Logistics domains, and slightly better in the DWR and Generic domains. This demonstrates (unsuprisingly) learning on conjunctive implications is insufficient to reason about effect of actions.

\subsubsection{Can Roberta reason about effect of actions?}
Table \ref{tab:maintable} summarizes the results on how Roberta finetuned on simpler worlds generalizes to out-of-domain  evaluation on larger more complex worlds. We can observe for \textit{Verify} questions, which is the same setting for RuleTakers, it can generalize to novel worlds to some extent. In Table \ref{tab:maintable}, all finetuned models are evaluated on N5, which is the highest complexity, but trained only on N1 and N3. For context, N1 for Blocks World is only 2 towers and N5 is 7 towers, and the world complexity for DWR and Logistics are even more complex. These results indicate that a transformer-encoder based architecture can reason about effect of actions, but there is still scope for further improvements, especially on more  complex real-world domains such as DWR. 

\subsubsection{How does explicit rules affect learning?}
From Figures  \ref{fig:knowtrend} and \ref{fig:noknowtrend},  and Table \ref{tab:maintable} we can observe the  effect of action-effect axioms in the three domains. On the novel world complexity OOD test sets, explicit rules show a conclusive positive trend only on the simplest of the domains, i.e, Blocks world. But on the action domain complexity OOD test  sets, explicit rules show a more profound effect especially when trained on simple  complexity questions (A1-2).
For DWR the implicit rules learned generalize better to novel worlds, with a smaller drop in accuracy compared with explicit rules. For Logistics the model trained on smallest complexity performs better with explicit rules. We also observe an interesting phenomenon: If the complexity is more than N2 (3 airplanes, 3 trucks, and 3 packages), the model achieves near perfect results, i.e, it learns to generalize to novel complex worlds ($>$3 for all). Our hypothesis is that the reasoning needed to answer \textit{Verify} questions may be sufficiently learned on that complexity, unlike Blocks World, or DWR. The Generic domain by definition needs action-effect axioms, and hence cannot be evaluated  without it. A model trained on lower complexity domains (lower number of fluents and actions) generalizes well to higher complexity domains (higher number of fluents  and  actions) but its performance is slightly worse than the performance of its counterpart model trained on higher complexity domains. The model is able to generalize to some extent even with a small number of examples (better than random).

In Figure \ref{fig:learningcurve} we observe the positive effect with lower number of samples for Blocks world and Logistics, but a different trend for DWR. Our hypothesis is that some world axioms are simpler to express in natural language, and rules for such domains can show a positive effect in reasoning accuracy. Exploring different ways to express complex domains in natural language and observing its effect on reasoning accuracy will be an interesting future direction.


\subsubsection{How many samples are needed to achieve a decent performance?} Figure \ref{fig:learningcurve} shows that in all the domains the model approaches its best performance on their in-distribution data around $10^4$ samples. Even though the rules augmented model's accuracy on the DWR domain appears to improve with increasing training samples till $5*10^4$ instances, it does not improve further. It is interesting to observe that other than the Logistics domain, in all other domains the model is able to achieve better than random accuracy (10\%+) with a small number of samples. 

\subsubsection{Can learning on Generic Domain transfer to BLD?}
We evaluate our models trained on the Generic domain on corresponding mapped questions from  BLD domains to verify, if learning about reasoning from actions from a generic domain of actions and fluents transfer to other worlds with complex action spaces. Table \ref{tab:maintable} summarizes the results. We observe that they do generalize to some extent, statistically significantly above random performance, even when trained on simpler action-fluent complexity. This is interesting to observe and shows the scope of defining and training neuro-symbolic models trained on knowledge representation abstractions to generalize to OOD domains. We evaluate only N1 complexity worlds as the textual description of initial state for N2 and beyond exceeds the 512 token limit of Roberta. The description increases  with complexity as the number of unique fluents and actions increase with increase in world complexity. Evaluating on transformers that enable greater than 512 tokens, such as Longformer \cite{beltagy2020longformer}, will be an interesting future work.

\subsubsection{What happens if we increase the length of action sequence?} Table \ref{tab:typeoodaction} and Figures  \ref{fig:knowtrend},\ref{fig:noknowtrend} show OOD generaliztion accuracy on different question types, with varying length of action sequences. We observe compared to novel worlds, the action depth complexity OOD generalization is a challenging task. This is expected as the hypothesis space or number of states to track increase with each action taken. One may note that the model does not have an explicit state space where the effect of each individual action is updated, stored and available for reference. Interestingly, models with explicit action-effect rules consistently outperform models that learn implicit rules from examples, especially  when trained on less complex worlds. Another observation is that learning on only 1/3 depth action sequence questions, the model is able to generalize to the depth of 5 action sequences on \textit{counting} questions. The counting questions need the model to count the number of fluents that are true in that state. But on the Generic domain, the model's performance is significantly poor, indicating the counting task to be harder. The models comparatively struggle to generalize on the \textit{Others} questions  in Blocks world and DWR domains. 



\begin{table}[t]
\centering
\resizebox{0.8\linewidth}{!}{
\begin{tabular}{@{}lccc@{}}
\toprule
\textbf{Model} & \textbf{Blocks World $\uparrow$} & \textbf{Logistics $\uparrow$} & \textbf{DWR $\uparrow$} \\ \midrule
A3~+~Rules & 81.0 & 81.0 & 75.0 \\
A3~+~NoRules & 80.4 & 78.0 & 77.0 \\
A5~+~Rules & \textbf{84.6} & \textbf{86.0} & \textbf{78.0} \\
A5~+~NoRules & 83.8 & 82.0 & \textbf{78.0} \\
\bottomrule
\end{tabular}}
\caption{Accuracy on the human authored paraphrased initial states and natural questions in the BLD domains (Overall).}
\label{tab:overtable}
\end{table}

\subsubsection{How does model generalize to linguistic variations in initial state and questions?} We asked expert human annotators to hand author 300 questions of the three types and world initial states for the BLD domains. These questions are used as test sets to evaluate zero-shot transfer on natural paraphrased questions. Its interesting to observe in Table \ref{tab:overtable} that Roberta is robust to paraphrased and natural questions, including where the objects are different from what it has seen during training (color blocks, to blocks re-named using numbers, animal names).


\section{Discussion}
A major advantage of our method to create synthetic data using ASP is that we can generate datasets of any size to facilitate learning of complex data-hungry neural models. As we generate answers from ASP solvers, we do not require costly and carefully designed human annotation frameworks to collect answers. Indeed our data generation method suffers with a lack of linguistic diversity, but recent state-of-the-art methods on paraphrasing, and other data augmentation techniques such as back-translation can be used to mitigate that. Moreover similar to the recent development in self-supervised learning techniques, with a shift away from a dataset focussed research, we propose methods to learn from synthetic instances and only evaluating on hand curated real-world representative domains, which avoids the pitfalls of linguistic priors and annotation bias.

There are several advantages of a natural language interface to reason about effect of actions. The requirement of carefully crafted ASP rules to define a domain needs a certain level of  expertise in knowledge representation and reasoning. This requirement is avoided when a user can describe a domain in natural language, define the actions and effects,  and then query about different states. Learning to perform reasoning about effects of actions from the Generic domain enables generalizing to any novel action domain. 

Although we only evaluate Roberta, contrary to previous results shown by \citeauthor{clark2020transformers}, we observe Roberta cannot perfectly answer and generalize to OOD worlds in both novel world complexity and action world complexity. Although ASP solvers with perfect rule definitions can achieve 100\% accuracy, but there are two bottlenecks: perfect semantic parsing from text to get the accurate initial state and human involvement in writing perfect rules. 
Our work is a step towards bridging this gap between these two approaches, by using the expertise of a symbolic ``reasoning about action'' solver to teach a robust neural model.  


\section{Related Work and Conclusion}

Earlier in Section~\ref{intro} we mentioned the recent work \cite{clark2020transformers} that motivated our research in this paper and also mentioned some of the prominent works related to reasoning about actions and change. In the context of answering questions with respect to natural language text where rules are explicitly given in natural language the main prior work is \cite{clark2020transformers}. That work in its extended version \cite{clark2020transformers2} mentions Task 15 of bAbI \cite{weston2016towards}, conditional probes in \cite{richardson2020probing}, and QuaRTz \cite{tafjord2019quartz} among examples of other work where application of general rules given in natural langauge is involved. Some of the tasks in \cite{weston2016towards} and the task in \cite{mishra2018tracking} involve reasoning about actions, but in them the effect of actions are {\em not explicitly stated in natural language}. In \cite{mitra2016addressing} explict  answer set programming (ASP) rules are learned from the bAbI  dataset and then used to reason using an ASP solver. In this paper we consider both the explicit and implicit cases, i.e., (i) where the effect of actions are explicitly given in natural language, similar to the approach in \cite{clark2020transformers}; and (ii) where the effect of actions are not explicitly given but a transformer has to implicitly reason with it. 
As mentioned in \cite{clark2020transformers2} transformers have been shown to be able to (learn to) emulate algorithms in various tasks such as semantic parsing \cite{he2019establishing}, machine translation \cite{wang2019learning}, symbolic integration \cite{lample2019deep} and mathematics \cite{saxton2018analysing}. Our work in this paper is perhaps the first work where transformers are tested with respect to a deeper knowledge representation and reasoning challenge, the challenge of reasoning about actions and their effects. In this regard it may be noted that, although reasoning about actions was formally introduced in 1963, it took decades of research before the associated ``frame problem'' was solved and human researchers were able to come up with formulations, as in \cite{reiter1991frame,gelfond1993representing}, that could reason about actions in a systematic and provable way. 


This domain of reasoning about actions and change is the focus of this paper and we show that when trained in a specific domain, transformers are able to reason about the effect of actions in that domain and within that complexity with high accuracy (90-98\%).
However, when testing with higher complexity the accuracy drops to 68-90\%. When trained in the generic domain the transfer learning accuracy with respect to the BLD domains is 57-83\%. Thus, even for the simplest aspect of RAC, which is reasoning about effects of actions, further research is needed for better transfer learning / out-of-domain accuracy. Beyond that there are harder challenges in RAC such as planning \cite{lifschitz1987semantics}, explanation, diagnosis \cite{baral2000formulating} and narrative reasoning \cite{shanahan1997solving,mueller2014commonsense}. 


\bibstyle{aaai21}
\bibliography{ref}

\section{Open Source}
We will make both our code and hand-authored samples public at https://xxx.github.com. As we procedurally generate the data, the synthetic data generation templates and ASP code is as follows. We also provide the detailed experimental setup, and experimental results for our plots. 

\section{Question-Answer Examples}
In this section, we provide the templates for four types of questions and training examples for each action domain (Blocks World, Logistics, Dock-Workers Robots and Generic Domain). 


Table \ref{tab:templates-blocks}, \ref{tab:templates-logistics}, \ref{tab:templates-dwr}, \ref{tab:templates-generic} shows a few question templates for the blocks world, logistics, dwr and generic domain respectively. Strings within the  \{.\} are variables and replaced during data generation. We create multiple paraphrased version of each template so that the model do not overfit to similar patterns. We chose randomly and select one question template from each template class.

\begin{table}[h]   
\centering
\resizebox{\linewidth}{!}{
\begin{tabular}{@{}ll@{}} 

\toprule
\textbf{Type} & \textbf{Templates} \\ \midrule
Verify &  Will the top of the \{block\} be empty \\
& after the \{block1\} is moved to the top of the \{block2\} ?\\
& Will the \{block\} be directly on the table \\
& after the \{block1\} is moved to the top of the \{block2\} block ?\\
& Would all the blocks be directly on the table \\
& after the \{block1\}  is moved to the top of the \{block2\}?\\
\hline
Counting &  How many blocks would be directly on the table \\
& after the \{block1\} is moved to the top of the \{block2\} ?\\
& Find the height of the tallest stack of blocks.\\
\hline
Others &  \{ name \} moved the \{ block\_being\_moved \} block to the top \\
    & of the \{block\_moved\_to\} block. Which block/blocks is/are above the\\ & \{block\_in\_question\} block ?\\
    & Which block is in between the \{top\_block\} block and the \{bottom\_block\}\\ 
    &block if \{name\} move  the \{block\_being\_moved\} block to the top of the \\
    & \{block\_moved\_to\} block ?\\

\bottomrule
\end{tabular}}
\caption{Sample templates of questions for Blocks world domain.}
\label{tab:templates-blocks}
\end{table}


\begin{table}[h]
\centering
\resizebox{\linewidth}{!}{
\begin{tabular}{@{}ll@{}}

\toprule
\textbf{Type} & \textbf{Templates} \\ \midrule
Verify &  Can the truck \{truck\} be moved from the\\
& location \{location\_from\} to the location \{location\_to\}?\\
& Can the airplane \{airplane\} be flown from the\\
&  location \{location\_from\} to the location \{location\_to\}?\\
& Can the airplane \{airplane\} be loaded with \\
& the object \{load\_object\} ?\\
& Can the airplane \{airplane\} be unloaded ?\\
\hline

Counting &  How many objects are present in the location \{loc\} ?\\
& How many trucks are present in the location \{loc\} ?\\
\hline

Others &  What is the location of the truck \{truck\} now ?\\
& Which trucks are present in the location \{loc\} ?\\

\bottomrule
\end{tabular}}
\caption{Templates of question for Logistics domain.}
\label{tab:templates-logistics}
\end{table}


\begin{table}[h]
\centering
\resizebox{\linewidth}{!}{
\begin{tabular}{@{}ll@{}}

\toprule
\textbf{Type} & \textbf{Templates} \\ \midrule
Verify &  Is \{crane\} empty?Is \{robot\} at \{location\}?\\
& Can \{container\} be loaded on \{robot\} in the current state?\\
& Can \{container\} be picked up by \{crane\} in the current state?\\
& Is the \{container\} at the top of \{pile\}?\\
& Is \{robot\} loaded with \{container\}?\\
& Is \{robot\} unloaded?\\
\hline

Counting &  how many cranes are empty?\\
& how many crane are not empty?\\
& how many containers are in \{pile\}?\\
& how many locations are occupied?\\
& how many robots are unloaded?\\
& how many robots are loaded?\\
\hline

Others &  which location does the \{robot\} at?\\
& \{robot\} is at which location? \\
& which pile does the \{container\} in?\\
& \{container\} is in which pile? \\
& which container is on top of \{pile\}?\\

\bottomrule
\end{tabular}}
\caption{Templates of question for DWR domain.}
\label{tab:templates-dwr}
\end{table}


\begin{table}[h]
\centering
\resizebox{\linewidth}{!}{
\begin{tabular}{@{}ll@{}} 

\toprule
\textbf{Type} & \textbf{Templates} \\ \midrule
Verify &  Can action \{action\} be executed in the initial state?\\
& Will the fluent \{fluent\} be true \\
& after executing action \{action\}?\\
& Will the fluent \{fluent\} be true \\
& after executing action \{action\} and then action \{action\_2\}?\\
\hline

Counting &  How many fluents would be true \\
& after executing action\{action\}?\\
& How many fluents would be true \\
& after executing action \{action1\} and then action \{action2\}?\\
& Would all the fluents be true after executing   \{action\}?\\

\bottomrule
\end{tabular}}
\caption{Templates of question for Generic domain.}
\label{tab:templates-generic}
\end{table}

The following are examples for each of the generated domains. The first example is from the \textbf{DWR domain}: \newline 
\hrule
\smallskip
\footnotesize{
\textbf{Initial State}:  These are locations: fishery, airfield.  Robots are: robot-3, r10.  Crane are: crane-9, crane-7.  There are piles: pile-9, pile-12.  These are containers: seashell, moccasin.  fishery is adjacent to airfield.  fishery has the pile pile-9. pile-12 is at airfield.  crane-9 is located at fishery. airfield has crane-7.  robot-3 presents at airfield.  fishery has r10.    robot-3 is unloaded.  r10 can hold a container.  crane-9 can hold a container.  crane-7 can hold a container.  These are stacked in order top to bottom : seashell, moccasin.  pile-12 has seashell at the top.\\ 
\textbf{Rules}: A robot can be at one location at a time. 
A crane can move a container from the top of a pile to an empty robot or to the top of another pile at the same location.
A container can be stacked in some pile on top of the pallet or some other container, loaded on a robot, or held by a crane.
A pile is a fixed area attached to a single location.
A crane belongs to a single location; it can manipulate containers within that location, between piles and robots.
Each robot can carry one container at a time.
Robots can move to free adjacent location.
A crane is empty if it is not holding container.
A robot is unloaded if it is not loaded with container.\\
\textbf{Action}: Crane-7 picks up seashell and seashell is loaded on r10. \\
\textbf{Verify} \\
\textit{Is r10 loaded with moccasin?}\;\;Ans: \textbf{No}.\\
\textit{Is moccasin on top of pile-12?}\;\;Ans: \textbf{Yes}.\\
\textbf{Counting} \\
\textit{how many robots are unloaded?} Ans: \textbf{1}.\\
\textit{how many cranes are not empty?} Ans: \textbf{0}.\\
\textbf{Others}
\textit{which container is on top of pile-12?} Ans: \textbf{moccasin}.\\
\textit{which location does the r10 at?} Ans: \textbf{fishery}.
}
\hrule

\normalsize
\vspace{3pt}
Following is the example from the \textbf{Generic Action Fluents Domain}:
\newline
\hrule
\footnotesize{
\noindent
\textbf{Initial State}:  Initially true fluents: 7, 1, 6, 8. \\ 
\textbf{Rules}: Actions are 1, 4, 6, 8, 5, 3, 2, 7 and fluents are 1, 2, 6, 7, 4, 5, 3, 8. Action 6 causes -6 if -7. Action 2 causes -8. Action 4 causes -8. Action 8 causes 3 if 7, 8. Action 1 causes -1. Action 7 causes -8 if 1, -5. Action 3 causes 8. Action 5 causes 2. Precondition for action 4 is 3, for action 3 is 8, for action 2 is -8.\\
\textbf{Action}: action 3 is executed. \\
\textbf{Verify}~~Q:) \textit{Will the fluent 1 be true after executing action 6?}\;\;Ans: \textbf{Yes}.\\
\textbf{Counting}~~Q:) \textit{How many fluents would be true after executing action 6?} Ans: \textbf{4}.
}
\vspace{2pt}
\hrule
\vspace{10pt}
\normalsize

Following is an example from the \textbf{Logistics} domain: 
\vspace{2pt}
\hrule
\noindent
\footnotesize{\textbf{Initial State}:  
The cities in this world are: delhi, uttarpradesh, rajasthan, Jammu, kerala. The locations defined are: market, industry, ground, fishery, airfield, depot, seaport, forestry .The objects which are transported are:  Bricks, Mirror, Computers . 
These are trucks:  Ashok, Tata, General . Air cargo is flown by:  Dreamliner, Boeing, Cirrus . 
The locations that have airport are : market, seaport, industry, ground . delhi city has the following locations: market, industry . uttarpradesh city has the following locations: ground, fishery. rajasthan city has the following locations: airfield, depot. The following locations are present in Jammu : seaport.  The kerala city has the forestry location. The Bricks cargo is at market. The Mirror is present at industry. The Ashok is located at market. industry is where Tata is present. The location ground has the cargo Computers. ground is where General is present. The location market has the plane Dreamliner. The location seaport has the plane Boeing. The Cirrus plane is at industry.\\
\textbf{Rules}: Vehicles airplane, truck transport packages between locations present in city. Trucks deliver within city. Airplane between airports associated with a location. Object can be loaded or unloaded in vehicle if present at same location. Unloaded package remain at  vehicle location . Loading object cause object to be inside vehicle.Driving truck from its current location to another causes truck to change its location. Flying airplane from current airport to another causes airplane to change its airport.\\
\textbf{Action}: \\Hannah started from the location industry and drove the truck tata to the location market. \\
\textbf{Verify}\\
\textit{Can Hannah drive the truck tata from the location market to the location seaport ?}\;\;Ans: \textbf{No}.\\
\textit{Is it possible for Hannah to move the truck tata from the location market to the location industry ?}\;\;Ans: \textbf{Yes}.\\
\textit{Can the object mirror be kept in the truck tata ?}\;\;Ans: \textbf{No}.\\
\textit{Can Hannah unload the truck tata ?}\;\;Ans: \textbf{No}.\\
\textbf{Counting}\\ \textit{How many trucks present in location market?} Ans: \textbf{2}.\\\textit{How many objects present in location seaport?} Ans: \textbf{0}.\\
\textbf{Others}\\
\textit{What is the location of the truck tata now ?} Ans: \textbf{market}.\\
\textit{Which objects are present in the location seaport ?} Ans: \textbf{none}.\\
\textbf{Action}: \\
Hannah started from the location industry and flew the airplane cirrus to the location ground.\\
\textbf{Verify}\\
\textit{Is it possible for Hannah to fly the airplane cirrus from the location ground to the location seaport ?}\;\;Ans: \textbf{Yes}.\\
\textit{Is it possible for Hannah to fly the airplane cirrus from the location ground to the location fishery ?}\;\;Ans: \textbf{No}.\\
\textit{Is it possible for Hannah to load the airplane cirrus with the object computers?}\;\;Ans: \textbf{No}.\\
\textbf{Counting}\\
\textit{How many airplanes are present in the location forestry ?}\;\;Ans: \textbf{0}.\\
\textbf{Others}\\
\textit{Which objects are present in the location market ?}\;\;Ans: \textbf{bricks}.
}
\vspace{1pt}
\hrule
\normalsize
\section{Translation to Generic Domain}
We translate all the actions domains to the generic domain. Table \ref{tab:translation} shows examples of translated examples.

\begin{table*}[ht]
\centering
\resizebox{0.8\linewidth}{!}
{
\def\arraystretch{1.2}
\begin{tabular}{@{}lll@{}} 
    \toprule
    \textbf{Domain} & \textbf{Question} & \textbf{Translation} \\ 
    \midrule
    Blocks &  Will the top of the \{block\}\\
    &  be empty after the \{block\_being\_moved\} is \\
    & moved to the top of the \{block\_moved\_to\}? 
    & Will the fluent \{fluent\_in\_question\} be true \\
    \hline
    Logistics &  
    Can the truck \{truck\} be moved from the \\
    & location \{location\_from\} to the location \\
    & \{location\_to\}?  
    & Can action \{action\} be executed in the initial state?\\
    \hline
    DWR &  
    Can \{container\} be picked up by {crane} in the current state? 
    & Is it possible to execute action 
    \{action\} in the current state?\\
    \bottomrule
\end{tabular}
}
\caption{Examples of translated questions from BLD domains to generic domain.}
\label{tab:translation}
\end{table*}

\section{Training Details}

In this section, we describe the hyperparameters used in training. Depending on the types of questions used in the training time,  we present the results for two training sets: including verify questions only, and including three types of questions. We also present the hand written examples at the end of this section.

\subsection{Training Parameters}
We use RoBERTa-Large fine-tuned on RACE dataset as our base model. We train all our models with a learning rate of 1e-5 for 3 epochs. We adjust the per gpu training batch size and gradient accumulation accordingly to fit in our 4 Nvidia V100 16GB GPUs. We keep maximum sequence length of 512 for our experiments to ensure that the whole context is used by the model. We adjust the save steps and logging steps for each experiments accordingly.

\subsection {Training with Verify Questions}

Based on the novel world complexity, we define five levels, and for each level, we generate 50K verify questions for training and testing seperately. To avoid bias, the size of negative and positive examples are equivalent. We train Roberta with each dataset (with or withour domain knowledge), and test the model on all five dataset (with or withour domain knowledge). The dataset with different complexity level as used in the training time is considered as OOD domain. Table \ref{tab:bw-tf},\ref{tab:log-tf},\ref{tab:dwr-tf} show the results without domain knowledge in training for Blocks World, logistic, and DWR seperately. Table \ref{tab:bw-tf-kn},\ref{tab:log-tf-kn},\ref{tab:dwr-tf-kn},\ref{tab:gen-tf-kn} show the results without domain knowledge in training for Blocks World, logistic, DWR, and generic seperately. Let Ni represent dataset of complexity level i. 

\begin{table}[t]
\centering
\resizebox{\linewidth}{!}{
\begin{tabular}{@{}cccccc@{}}
\toprule
\textbf{Train \slash Test} & N1 & N2 & 

N3 & N4  & N5 \\ \midrule
N1 &99.98 & 96.73 & 91.64 & 84.53 & 83.03 \\
N2 &99.91 &99.87 & 93.31 & 80.36 & 79.42 \\
N3 & 99.88 & 99.76 &99.69 & 86.78 & 82.25 \\
N4 & 99.85 & 99.68 & 99.65 &99.99 & 83.24 \\
N5 & 99.83 & 99.73 & 99.48 &99.96 & 99.82 \\ 
\bottomrule
\end{tabular}
}
\caption{Accuracy of Blocks Worlds domain on verify questions on five novel world test scenarios without knowledge.}
\label{tab:bw-tf}
\end{table}

\begin{table}[t]
\centering
\resizebox{\linewidth}{!}{
\begin{tabular}{@{}cccccc@{}}
\toprule
\textbf{Train \slash Test} & N1 & N2 & 

N3 & N4  & N5 \\ \midrule
N1 & 99.11 & 90.02 &89.74 &86.25 & 85.28 \\
N2 &99.36 &99.34 & 92.78 & 88.36 & 88.27 \\
N3 & 99.49 & 99.07 &99.04 & 91.98 &90.92 \\
N4 & 98.39 & 99.32 & 98.96 & 99.67 & 90.88 \\
N5 & 99.75 & 99.13 & 99.12 & 99.69 & 99.72 \\ 
\bottomrule
\end{tabular}
}
\caption{Accuracy of Blocks Worlds domain on verify questions on five novel world test scenarios with knowledge.}
\label{tab:bw-tf-kn}
\end{table}

\begin{table}[t]
\centering
\resizebox{\linewidth}{!}{
\begin{tabular}{@{}cccccc@{}}
\toprule
\textbf{Train \slash Test} & N1 & N2 & 
N3 & N4  & N5 \\ \midrule
N1 & 97.47 & 89.21 &80.25 &77.36 & 69.89 \\
N2 &99.98 &100 &99.99 & 99.98 & 99.97 \\
N3 & 99.75 & 99.90 &99.90 & 99.96&99.97 \\
N4 & 99.91 & 99.96 & 99.98 & 99.99 & 100 \\
N5 & 99.86 & 99.86 & 99.89 & 99.97 & 99.99 \\ 
\bottomrule
\end{tabular}
}
\caption{Accuracy of Logistic domain on verify questions on five novel world test scenarios without knowledge.}
\label{tab:log-tf}
\end{table}

\begin{table}[t]
\centering
\resizebox{\linewidth}{!}{
\begin{tabular}{@{}cccccc@{}}
\toprule
\textbf{Train \slash Test} & N1 & N2 & 

N3 & N4  & N5 \\ \midrule
N1 &99.99 & 93.98 & 87.33 &78.31 & 71.31 \\
N2 &96.98 &99.97 & 99.97 &99.93 & 99.91 \\
N3 & 99.46 & 99.99 &99.98& 99.99 & 100 \\
N4 & 99.95 &99.96 & 99.99&99.99&99.99 \\
N5 &99.91&99.94&99.94 &99.96 & 99.99 \\ 
\bottomrule
\end{tabular}
}
\caption{Accuracy of Logistic domain on verify questions on five novel world test scenarios with knowledge.}
\label{tab:log-tf-kn}
\end{table}

\begin{table}[t]
\centering
\resizebox{\linewidth}{!}{
\begin{tabular}{@{}cccccc@{}}
\toprule
\textbf{Train \slash Test} & N1 & N2 & 

N3 & N4  & N5 \\ \midrule
N1 & 83.22&81.38& 80.32 & 79.05 & 76.19 \\
N2 &80.34 &79.06 & 78.52& 76.51 & 76.07 \\
N3 & 83.45 & 83.76 &79.28& 80.37 & 77.40 \\
N4 & 68.85 &76.16 &71.96 &75.67 & 71.37 \\
N5 & 77.20 &77.41& 76.97 &76.78 & 76.46\\ 
\bottomrule
\end{tabular}
}
\caption{Accuracy of DWR domain on verify questions on five novel world test scenarios without knowledge.}
\label{tab:dwr-tf}
\end{table}

\begin{table}[t]
\centering
\resizebox{\linewidth}{!}{
\begin{tabular}{@{}cccccc@{}}
\toprule
\textbf{Train \slash Test} & N1 & N2 & 
N3 & N4  & N5 \\ \midrule
N1 & 85.68&82.65& 80.47 & 79.75 & 72.05 \\
N2 &82.77 &82.28 & 79.93& 79.65 & 71.99 \\
N3 & 81.24 & 81.51 &79.17& 79.77 & 72.96 \\
N4 & 77.06 &77.02 &74.92 &76.52 & 70.30 \\
N5 &78.36 &77.41& 75.35 &76.46 & 71.11\\ 
\bottomrule
\end{tabular}
}
\caption{Accuracy of DWR domain on verify questions on five novel world test scenarios with knowledge.}
\label{tab:dwr-tf-kn}
\end{table}

\begin{table}[t]
\centering
\resizebox{\linewidth}{!}{
\begin{tabular}{@{}cccccc@{}}
\toprule
\textbf{Train \slash Test} & N1 & N2 & N3 & N4  & N5 \\ \midrule
N1 & 79.03 & 74.97 & 71.61 & 79.70 & 78.46 \\
N2 &78.76 &79.86 & 78.38 & 80.63 & 77.40 \\
N3 & 78.97 &79.21 &84.90 & 83.70 & 81.27 \\
N4 & 78.74 & 79.86 & 83.78 & 85.72 & 82.93 \\
N5 & 78.36 & 79.75 & 84.24 & 86.65 & 83.91\\ 
\bottomrule
\end{tabular}
}
\caption{Accuracy of Generic domain on verify questions on five novel world test scenarios with knowledge.}
\label{tab:gen-tf-kn}
\end{table}

\subsection {Training with Three Types of Questions}

Based on the depth complexity, we define five levels. Except for generic domain,  we generate 40K verifty questions, 30K Counting questions and 30K others questions for training and 50K of three types for testing seperately for other three domains. We train Roberta with each dataset(with or withour domain knowledge), and test the model on all five dataset (with or withour domain knowledge). The dataset with different complexity level as used in the training time is considered as OOD domain.Table show the results without domain knowledge in training for Blocks World, logistic, DWR, and generic seperately.  Table \ref{tab:bw-three},\ref{tab:log-three},\ref{tab:dwr-three} show the results without domain knowledge in training for Blocks World, logistic and DWR seperately. Table \ref{tab:bw-three-kn},\ref{tab:log-three-kn},\ref{tab:dwr-three-kn} show the results without domain knowledge in training for Blocks World, logistic and DWR generic seperately. Let Ai represent dataset of complexity level i. 

\begin{table*}[t]
\centering
\small
\resizebox{\linewidth}{!}{
\begin{tabular}{@{}lccccccccccccccc@{}}
\toprule
 & \multicolumn{3}{c}{A1} & \multicolumn{3}{c}{A2} & \multicolumn{3}{c}{A3} & \multicolumn{3}{c}{A4} &
 \multicolumn{3}{c}{A5}\\
\textbf{Train \slash Test} & \textbf{T/F} & \textbf{Num} & \textbf{Other} & \textbf{T/F} & \textbf{Num} & \textbf{Other} & \textbf{T/F} & \textbf{Num} & \textbf{Other} & \textbf{T/F} & \textbf{Num} & \textbf{Other} & \textbf{T/F} & \textbf{Num} & \textbf{Other} \\ \midrule
A1 & 
93.94& 85.88& 53.80&
90.88& 84.40& 52.06&
87.08& 79.30& 48.24&
83.30& 75.19& 40.91&
76.31& 72.40& 38.61
\\
A2 &
93.40& 90.34& 59.64&
92.82&  87.95& 58.09&
91.54& 86.38& 54.92&
89.46& 85.19& 52.28&
87.12& 82.41& 51.48
\\
A3 &
93.20& 96.27& 65.84&
93.12& 94.62& 63.72&
92.62& 94.25& 62.28&
92.10& 92.40& 61.71&
90.81& 90.20& 59.77
\\
A4 &
95.54& 93.84& 64.61&
95.65& 93.02& 64.76&
95.16& 92.62& 63.81&
94.45& 91.71& 62.74&
94.32& 91.43& 60.80
\\
A5 &
95.1& 93.81& 68.81&
95.67& 92.86& 68.25&
95.75& 92.56& 68.13&
95.82& 92.33& 66.83&
95.2& 92.18& 66.48
\\ \bottomrule
\end{tabular}
}
\caption{Accuracy of Block World domain on three types of questions on five novel world test scenarios without knowledge.}
\label{tab:bw-three}
\end{table*}

\begin{table*}[t]
\centering
\small
\resizebox{\linewidth}{!}{
\begin{tabular}{@{}lccccccccccccccc@{}}
\toprule
 & \multicolumn{3}{c}{A1} & \multicolumn{3}{c}{A2} & \multicolumn{3}{c}{A3} & \multicolumn{3}{c}{A4} &
 \multicolumn{3}{c}{A5}\\
\textbf{Train \slash Test} & \textbf{T/F} & \textbf{Num} & \textbf{Other} & \textbf{T/F} & \textbf{Num} & \textbf{Other} & \textbf{T/F} & \textbf{Num} & \textbf{Other} & \textbf{T/F} & \textbf{Num} & \textbf{Other} & \textbf{T/F} & \textbf{Num} & \textbf{Other} \\ \midrule
A1 & 
97.42 & 98.73& 63.64 &
92.94& 88.35& 57.94 & 
88.20& 82.15&54.28& 
84.27& 77.89& 49.65 &
81.08& 75.13& 47.18
\\
A2 &
97.47& 94.24& 61.68 &
97.29&  92.76& 58.42 &
96.49& 90.66& 56.04 &
94.74& 87.78& 51.68 &
92.74& 85.34& 50.08 
\\
A3 &
96.94& 96.51& 65.63 &
96.8& 95.55& 63.52&
96.64& 95.51& 62.71&
95.85& 94.60& 60.88&
94.95& 92.94& 60.20
\\
A4 &
96.03& 96.36& 66.75&
96.13& 95.08& 64.74&
96.02& 94.52& 63.00&
95.58& 93.62& 62.76&
95.09& 92.68& 62.52
\\
A5 &
96.36& 94.53& 65.81&
96.43& 93.00& 64.09&
96.19& 92.70& 62.60&
95.91& 91.92& 60.86&
95.54& 91.66& 61.49
\\
\bottomrule
\end{tabular}
}
\caption{Accuracy of Blocks World domain on three types of questions on five novel world test scenarios with knowledge.}
\label{tab:bw-three-kn}
\end{table*}

\begin{table*}[t]
\centering
\small
\resizebox{\linewidth}{!}{
\begin{tabular}{@{}lccccccccccccccc@{}}
\toprule
 & \multicolumn{3}{c}{A1} & \multicolumn{3}{c}{A2} & \multicolumn{3}{c}{A3} & \multicolumn{3}{c}{A4} &
 \multicolumn{3}{c}{A5}\\
\textbf{Train \slash Test} & \textbf{T/F} & \textbf{Num} & \textbf{Other} & \textbf{T/F} & \textbf{Num} & \textbf{Other} & \textbf{T/F} & \textbf{Num} & \textbf{Other} & \textbf{T/F} & \textbf{Num} & \textbf{Other} & \textbf{T/F} & \textbf{Num} & \textbf{Other} \\ \midrule
A1 &
100&99.80& 100&
84.77&34.95& 59.49&
78.19&38.68&78.73&
75.86& 40.71& 71.31&
75.29&33.85& 68.12
\\
A2 & 
99.99& 98.79&100&
99.97& 99.22& 100&
100&98.07& 92.32&
99.76& 95.30& 96.06&
99.56& 91.58& 94.43
\\
A3 &
99.99& 98.98& 100&
100& 99.37& 100&
100& 99.44& 100&
99.65& 93.04& 99.68&
99.62& 90.48& 98.70
\\ 
A4 &
100& 98.66& 100&
100&,99.05& 100&
100&,99.66& 100&
100&98.41& 99.98&
99.96&95.83& 99.91
\\ 
A5 & 
99.97& 92.95& 99.99&
99.98&95.93& 100&
100&95.42& 100&
99.99& 94.93& 100&
99.96&91.58& 99.98
\\ \bottomrule
\end{tabular}
}
\caption{Accuracy of Logistic domain on three types of questions on five novel world test scenarios without knowledge.}
\label{tab:log-three}
\end{table*}

\begin{table*}[t]
\centering
\small
\resizebox{\linewidth}{!}{
\begin{tabular}{@{}lccccccccccccccc@{}}
\toprule
 & \multicolumn{3}{c}{A1} & \multicolumn{3}{c}{A2} & \multicolumn{3}{c}{A3} & \multicolumn{3}{c}{A4} &
 \multicolumn{3}{c}{A5}\\
\textbf{Train \slash Test} & \textbf{T/F} & \textbf{Num} & \textbf{Other} & \textbf{T/F} & \textbf{Num} & \textbf{Other} & \textbf{T/F} & \textbf{Num} & \textbf{Other} & \textbf{T/F} & \textbf{Num} & \textbf{Other} & \textbf{T/F} & \textbf{Num} & \textbf{Other} \\ \midrule
A1 &
100& 79.77& 100&
98.26& 37.34&62.26&
98.81& 59.75& 85.17&
97.36& 55.01& 77.46&
97.98&55.20& 75.43
\\
A2 &
100& 97.60& 100&
100& 99.21&100&
100& 97.57& 96.73&
99.95& 92.67& 98.40&
99.83& 88.79&97.14
\\
A3 &
98.85&70.64&65.92&
99.89& 62.95&99.87&
98.80& 76.28&99.98&
98.85& 68.59& 97.44&
98.07& 65.30&95.60 
\\
A4 &
100& 95.92& 100&
99.94&97.68& 100&
100& 97.11& 99.99&
99.95& 96.53&99.89&
100&91.73&99.78
\\ 
A5 & 
99.77& 81.55& 99.65&
99.88& 87.14&98.33&
99.94& 77.60& 98.84&
99.85& 79.25& 96.51&
99.42&75.91& 94.05
\\ \bottomrule
\end{tabular}
}
\caption{Accuracy of Logistic domain on three types of questions on five novel world test scenarios with knowledge.}
\label{tab:log-three-kn}

\end{table*}

\begin{table*}[t]
\centering
\small
\resizebox{\linewidth}{!}{
\begin{tabular}{@{}lccccccccccccccc@{}}
\toprule
 & \multicolumn{3}{c}{A1} & \multicolumn{3}{c}{A2} & \multicolumn{3}{c}{A3} & \multicolumn{3}{c}{A4} &
 \multicolumn{3}{c}{A5}\\
\textbf{Train \slash Test} & \textbf{T/F} & \textbf{Num} & \textbf{Other} & \textbf{T/F} & \textbf{Num} & \textbf{Other} & \textbf{T/F} & \textbf{Num} & \textbf{Other} & \textbf{T/F} & \textbf{Num} & \textbf{Other} & \textbf{T/F} & \textbf{Num} & \textbf{Other} \\ \midrule
A1 &
82.65&87.11&74.65&
82.08&78.85&72.77&
82.47&78.78&73.4&
75.54&77.64&72.96&
78.38&78.73&74.88
\\
A2 &
81.85&85.31&74.40&
82.06&83.02&72.64&
84.79&81.05&74.88&
78.67&80.51&71.49&
77.62&81.58&68.08
\\
A3 &
79.32&81.11&68.92&
79.58&78.55&67.42&
81.35&79.20&71.77&
80.09&76.46&68.59&
75.95&83.93&61.66
\\
A4 & 
76.22&78.38&38.79&
75.01&76.65&39.06&
78.59&77.50&37.79&
73.84&79.19&42.04&
76.87&85.00&35.12
\\ 
A5 & 
74.03&78.61&33.06&
73.93&76.36&33.55&
76.91&76.07&30.84&
72.03&78.66&27.64&
85.21&85.13&39.78
\\ \bottomrule
\end{tabular}
}
\caption{Accuracy of DWR domain on three types of questions on five novel world test scenarios without knowledge.}
\label{tab:dwr-three}
\end{table*}

\begin{table*}[t]
\centering
\small
\resizebox{\linewidth}{!}{
\begin{tabular}{@{}lccccccccccccccc@{}}
\toprule
 & \multicolumn{3}{c}{A1} & \multicolumn{3}{c}{A2} & \multicolumn{3}{c}{A3} & \multicolumn{3}{c}{A4} &
 \multicolumn{3}{c}{A5}\\
\textbf{Train \slash Test} & \textbf{T/F} & \textbf{Num} & \textbf{Other} & \textbf{T/F} & \textbf{Num} & \textbf{Other} & \textbf{T/F} & \textbf{Num} & \textbf{Other} & \textbf{T/F} & \textbf{Num} & \textbf{Other} & \textbf{T/F} & \textbf{Num} & \textbf{Other} \\ \midrule
A1 &
83.08&87.52&74.60&
81.62&81.96&72.34&
78.85&79.63&72.93&
77.12&77.94&73.67&
78.51&80.30&74.12
\\
A2 & 
82.17&86.93&20.25&
82.97&84.58&21.26&
86.47&82.04&21.5&
82.68&81.94&20.81&
80.54&81.8&30.77
\\
A3 &
78.16&81.38&71.46&
79.31&79.50&69.25&
81.21&78.93&75.48&
78.41&78.36&68.10&
76.84&87.51&74.14
\\
A4 & 
74.93&73.76&34.90&
76.06&69.35&35.46&
77.16&75.80&33.44&
73.63&71.88&36.03&
78.88&81.29&34.74
\\ 
A5 &
74.65&76.56&37.16&
75.80&72.75&38.96&
76.84&72.03&35.94&
77.05&72.80&35.90&
82.44&87.44&40.42
\\ \bottomrule
\end{tabular}
}
\caption{Accuracy of DWR domain on three types of questions on five novel world test scenarios with knowledge.}
\label{tab:dwr-three-kn}
\end{table*}

\subsection {Hand Written Examples}

To test the model's generality, for each domain, we manually create (around) 100 examples each for Blocks Worlds, Logistic and DWR domain. We present the examples for each domain as follows.

\subsubsection{Logistics}
We created adversarial samples from our synthetic action data for each of the question types by modifying them various adversarial settings as can be seen from Table \ref{tab:log_adv}. For Counting questions, we modified the answers by either adding or removing an entity belonging to any of object, truck or airplane about which the question have been asked. For ``verify" and other ``questions", we modified the entity names which have not been seen in either training or previous testing phases.

\noindent
\hrule
\vspace{1pt}
\footnotesize{
            \textbf{Initial State}: The cities in this world are:  dispur,nagaland,karnataka,gujarat . The locations defined are:  market,park,fishery,industry,seaport,forestry,ground . The objects which are transported are:  Bricks,Batteries,Cement . The Land cargo is driven by:  Maruti,Diesel,Ashok,\textit{\textbf{hover}} . These are planes:  Tejas,Cirrus . The airports are located at:  seaport,ground . The following locations are present in dispur : market,park . nagaland city has the following locations: fishery,industry,seaport,forestry . karnataka city has the following locations: ground . The Bricks cargo is at market. The Batteries is present at park. The Maruti is located at market. park is where Diesel is present. The location market has truck \textit{\textbf{hover}}. The location fishery has the cargo Cement. fishery is where Ashok is present. The Tejas plane is at seaport. The Cirrus is at hangar in ground.\\
            \textbf{Action}:
            The truck ashok is driven from the location fishery to the location industry.  The truck ashok is driven from the location industry to the location seaport.  Gabriel started from the location seaport and drove the truck ashok to the location forestry.  \\
            \textbf{Q:}How many trucks are present in the location market ? \textbf{Ans:} 2\\
            \textbf{Change:}New truck hover introduced at market location which changes the answer old answer to the same question to 2.
}
\vspace{1pt}
\hrule

\begin{table}[t]
\centering
\resizebox{\linewidth}{!}{
\begin{tabular}{llr}
\toprule
\textbf{Type} & \textbf{Adversarial Change} & \textbf{Count}  \\ \midrule
Counting & Entity added, count Increased & 18\\
Counting & Entity removed, count Decreased & 8\\
Verify & Random Words Changed & 20\\
Verify & All Question Entities Changed & 7\\
Verify & Question Paraphrased/Modified & 15\\
Verify & Main Question Object Changed & 5\\
Others & Main Question Word Changed & 7\\
Others & Randomly Changed Name  & 20\\
\bottomrule
\end{tabular}
}
\caption{Types of Manual samples created for Logistics Domain}
\label{tab:log_adv}
\end{table}

\normalsize
\subsubsection{Dock-Workers Robots}
We create hand written DWR samples by changing the locations, robots, and containers which have not been seen in the training time. We also modify the description of initial state by injecting natural language phrases. We only ask verify and Counting questions.
\noindent
\vspace{1pt}
\hrule
\footnotesize{
    \textbf{Initial State}:  We have three locations in university: teaching building, office and library .  Robots can move and we have two robots: Bob, Jame.  Three cranes are: Mary, Patricia, Jennifer.  These are piles: pile\_10, pile\_1, p4.  We have two containers: container1, container2.  Mary is located at teaching building. library has Patricia. Jennifer is located at office.  Bob presents at library.  teaching building has Jame.    Bob has no containers.  Jame has no load.  Mary can hold a container.  Patricia holds no container.  Jennifer holds no container.   container1 is top of pile\_1. container2 presents at the top of p4.  teaching building and library are adjacent. library is adjacent to office. pile\_10 is attached to teaching building. library has the pile pile\_1. office has the pile p4.\\ 
   \textbf{Rules}: A robot can be at one location at a time. 
    A crane can move a container from the top of a pile to an empty robot or to the top of another pile at the same location.
    A container can be stacked in some pile on top of the pallet or some other container, loaded on a robot, or held by a crane.
    A pile is a fixed area attached to a single location.
    A crane belongs to a single location; it can manipulate containers within that location, between piles and robots.
    Each robot can carry one container at a time.
    Robots can move to free adjacent location.
    A crane is empty if it is not holding container.
    A robot is unloaded if it is not loaded with container.\\
    \textbf{Action}: Patricia picks up container1. \\
    \textbf{Verify Q:)} 
    \textit{Is Patricia empty?}\;\;Ans: \textbf{No}.\\
    \textbf{Action}: Bob moves to library. \\
    \textbf{Counting Q:)} 
    \textit{how many containers are in pile\_1?}\;\;Ans: \textbf{1}.
}
\vspace{1pt}
\hrule
\normalsize

\subsubsection{Blocks World}
For Blocks world we asked annotators to modify the name of blocks, and tower descriptions. Following is an example:

\noindent
\hrule
\vspace{1pt}
\footnotesize{
    \textbf{Initial State}:  There are five blocks  on the table. The blocks are zebra, monkey, giraffe, lion and tiger. Three blocks are stacked in a tower. The blocks in this stack are, penguin, parrot and eagle.
   \textbf{Rules}: A clear block can be moved to the top of the table. 
    A clear block can be moved to the top of another clear block.
    Moving a block to the top of the table causes the block to be on the table.
    Moving a block to the top of another block causes the first block to be on top of the second block and makes the second block not clear.
    Moving a block causes it to be no longer at where it was earlier.
    Moving a block  that was on another block cauese the second block to be clear.\\
    \textbf{Action}: The eagle block is moved from the top of the stack and placed on top lion. \\
    \textbf{Verify Q:} 
    \textit{Is block Lion empty?}\;\;Ans: \textbf{No}.\\
    \textbf{Action}: The giraffe block is placed on top tiger block. \\
    \textbf{Count Q:} 
    \textit{How many height atleast 2 stacks are present?} Ans: \textbf{2}.
}
\vspace{1pt}
\hrule
\normalsize

\section{ASP Code}
In this section, we provide the ASP code for each action domain. 
\subsection{Blocks World}

\begin{lstlisting}[language=clingo,caption=A Blocks World Domain, label=lst:blocks, 
mathescape=true,xleftmargin=.04\textwidth, breaklines=true]
time(0..len). 

fluent(on(X,Y))    :- block(X),block(Y), X!=Y. 
fluent(ontable(X)):- block(X).
fluent(clear(X))    :- block(X).
fluent(holding(X)):- block(X).
fluent(handempty). 

action(pickup(X)):- block(X).
action(putdown(X)):- block(X).
action(stack(X,Y)):- block(X),block(Y),  X!=Y.
action(unstack(X,Y)):- block(X),block(Y), X!=Y.  

executable(pickup(X), T):- time(T), action(pickup(X)), holds(handempty, T), holds(clear(X), T), holds(ontable(X), T).
executable(putdown(X), T):- time(T), action(putdown(X)), holds(holding(X), T).
executable(stack(X,Y), T):- time(T), action(stack(X,Y)), holds(holding(X), T), holds(clear(Y), T).
executable(unstack(X,Y), T):- time(T), action(unstack(X, Y)), holds(handempty, T), holds(on(X,Y), T), holds(clear(X), T).

holds(neg(ontable(X)), T+1):- time(T),occ(pickup(X),T).
holds(neg(handempty), T+1):- time(T),occ(pickup(X),T).
holds(holding(X), T+1):- time(T),occ(pickup(X),T).

holds(neg(holding(X)), T+1):- time(T),occ(putdown(X),T).
holds(clear(X), T+1):- time(T),occ(putdown(X),T).
holds(ontable(X), T+1):- time(T),occ(putdown(X),T).
holds(handempty, T+1):- time(T),occ(putdown(X),T).

holds(clear(X), T+1):- time(T),occ(stack(X,Y),T).
holds(neg(clear(Y)), T+1):- time(T),occ(stack(X,Y),T).
holds(neg(holding(X)), T+1):- time(T),occ(stack(X,Y),T).
holds(handempty, T+1):- time(T),occ(stack(X,Y),T).
holds(on(X,Y), T+1):- time(T),occ(stack(X,Y),T).

holds(neg(clear(X)), T+1):- time(T),occ(unstack(X,Y),T).
holds(clear(Y), T+1):- time(T),occ(unstack(X,Y),T).
holds(holding(X), T+1):- time(T),occ(unstack(X,Y),T).
holds(neg(handempty), T+1):- time(T),occ(unstack(X,Y),T).
holds(neg(on(X,Y)), T+1):- time(T),occ(unstack(X,Y),T).

holds(F, 0) :- initially(F). 
holds(neg(F), 0):- fluent(F), not initially(F). 

holds(F, T+1) :- time(T), fluent(F), holds(F, T), not holds(neg(F), T+1), T < len. 
holds(neg(F), T+1) :- time(T), fluent(F), holds(neg(F), T), not holds(F, T+1), T < len. 

1{occ(A,T) : action(A)} 1 :- time(T), T < len. 

:- time(T), action(A), occ(A, T), not executable(A, T). 
\end{lstlisting}

\subsection{Logistics}

\begin{lstlisting}[language=clingo,caption= Logistics Domain, label=lst:logistic, 
mathescape=true,xleftmargin=.04\textwidth, breaklines=true]
time(0..len). 

fluent(at(O, L)):- object(O), location(L). 
fluent(at(O, L)):- truck(O), location(L). 
fluent(at(O, L)):- airplane(O), location(L). 

fluent(in(O, T)):- object(O), truck(T). 
fluent(in(O, A)):- object(O), airplane(A). 
fluent(outside(O)):- object(O).   
 
action(load_truck(O, T, L))           :- object(O), truck(T), location(L).
action(load_airplane(O, A, L))     :- object(O), airplane(A), location(L).
action(unload_truck(O, T, L))       :- object(O), truck(T), location(L).
action(unload_airplane(O, A, L))  :- object(O), airplane(A), location(L).
action(drive_truck(T, L1, L2, C))   :- truck(T), location(L1), location(L2),  L1!=L2, city(C), in_city(L1, C), in_city(L2, C).
action(fly_airplane(A, L1, L2))      :- airplane(A), airport(L1), airport(L2), L1!=L2.

% load if at the same location 
executable(load_truck(O, Tr, L), T)          :- time(T), action(load_truck(O, Tr, L)), holds(at(O,L),T), holds(at(Tr,L),T), holds(outside(O), T).
executable(load_airplane(O, A, L), T)     :- time(T), action(load_airplane(O,A,L)),  holds(at(O,L),T), holds(at(A,L),T), holds(outside(O), T).

%% unload an object if the object is inside the truck or the airplane 
executable(unload_truck(O, Tr, L), T)     :- time(T), action(unload_truck(O, Tr, L)), holds(at(Tr,L),T), holds(in(O,Tr), T).
executable(unload_airplane(O, A, L), T) :- time(T), action(unload_airplane(O, A, L)), holds(at(A,L),T), holds(in(O,A), T).

%% drive or fly requires being at the starting location 
executable(drive_truck(Tr, L1, L2, C), T)  :- time(T), action(drive_truck(Tr, L1, L2, C)), holds(at(Tr,L1), T).
executable(fly_airplane(A, L1, L2), T)     :- time(T), action(fly_airplane(A, L1, L2)),  holds(at(A,L1),T).

%% load an object into a truck
holds(in(P,Tr), T+1)        :- time(T),occ(load_truck(P,Tr,L),T), holds(at(P,L), T).

%% load an object into an airplane 
holds(in(P,A), T+1)         :- time(T),occ(load_airplane(P,A,L),T).

%% unload an object from a truck
holds(neg(in(P,Tr)), T+1):- time(T),occ(unload_truck(P,Tr,L),T).
holds(outside(P), T+1):- time(T),occ(unload_truck(P,Tr,L),T).

%% unload an object from an airplane
holds(neg(in(P,A)), T+1):- time(T),occ(unload_airplane(P,A,L),T).
holds(outside(P), T+1):- time(T),occ(unload_airplane(P,A,L),T).

%% driving from one place to another  
holds(at(Tr,L1), T+1):- time(T),occ(drive_truck(Tr,L,L1,C),T).

%% flying from one airport to another 
holds(at(A,L1), T+1):- time(T),occ(fly_airplane(A,L,L1),T).

%% static causal laws 
holds(neg(at(P,L1)), T):- time(T), fluent(at(P,L)), fluent(at(P,L1)), holds(at(P,L), T), L != L1.  
holds(at(P,L1), T) :- time(T), holds(at(Tr,L1), T), holds(in(P,Tr), T). 
holds(at(P,L1), T) :- time(T), holds(at(A,L1), T),  holds(in(P,A), T). 
holds(neg(outside(P)), T) :- time(T), holds(in(P,Tr), T). 
holds(neg(outside(P)), T) :- time(T), holds(in(P,A), T). 

%% initial condition 
holds(F, 0) :- initially(F). 
holds(neg(F), 0):- fluent(F), not initially(F). 

%% inertial 
holds(F, T+1) :- time(T), fluent(F), holds(F, T), not holds(neg(F), T+1), T < len. 
holds(neg(F), T+1) :- time(T), fluent(F), holds(neg(F), T), not holds(F, T+1), T < len. 

1{occ(A,T) : action(A)} 1 :- time(T), T < len. 
:- time(T), action(A), occ(A, T), not executable(A, T). 

\end{lstlisting}

\subsection{Dock-Workers Robots}

\begin{lstlisting}[language=clingo,caption= Dock-Workers Robots Domain, label=lst:dwr, 
mathescape=true,xleftmargin=.04\textwidth, breaklines=true]
time(0..len). 

%% Fluents 
fluent(occupied(L)):-  location(L). 
fluent(at(R, L)):-  robot(R), location(L). 
fluent(loaded(R, C)):-  robot(R), container(C). 
fluent(unloaded(R)):-  robot(R). 
fluent(holding(K, C)):-  crane(K), container(C). 
fluent(empty(K)):-  crane(K). 
fluent(in(C, P)):-  container(C), pile(P). 
fluent(on(C1, C2)):-  container(C1), container(C2), C1 != C2. 
fluent(top(C, P)):-  container(C), pile(P). 

%% static causal laws 

%% location of robot 
holds(occupied(L),  T)         :- time(T), fluent(at(R, L)), holds(at(R, L), T).  
holds(neg(at(R, L)), T)         :- time(T), fluent(at(R, L)), fluent(at(R, L1)), holds(at(R, L1), T), L != L1.  

%% unloaded robot 
holds(neg(unloaded(R)), T)  :-  time(T), fluent(loaded(R, C)), holds(loaded(R, C), T). 
holds(unloaded(R), T)          :-  time(T), fluent(unloaded(R)), not holds(neg(unloaded(R)), T). 

%% empty crane
holds(neg(empty(K)), T)       :-  time(T), fluent(holding(K, C)), holds(holding(K, C), T). 
holds(empty(K), T)               :-  time(T), fluent(empty(K)), not holds(neg(empty(K)), T). 

%% not being on top of a pile if some crane is holding it, some other container is on it, or a robot is loading with it
holds(neg(top(C, P)), T)       :-  time(T), fluent(holding(K, C)), fluent(top(C, P)), holds(holding(K, C), T). 
holds(neg(top(C, P)), T)       :-  time(T), fluent(loaded(R, C)), fluent(top(C, P)), holds(loaded(R, C), T). 
holds(neg(top(C', P)), T)      :-  time(T), fluent(top(C', P)), fluent(on(C, C')), holds(on(C, C'), T). 

%% a container is in a pile if it is below some container in the pile or on top of the pile 
holds(in(C, P), T)                  :-  time(T), fluent(on(C', C)), fluent(in(C, P)), holds(on(C',C), T), holds(in(C', P), T).  
holds(in(C, P), T)                  :-  time(T), fluent(top(C,P)), holds(top(C, P), T).  

%% a container is not in a pile if in another pile or some crane is holding it or some robot is loading with it  
holds(neg(in(C, P)), T)         :-  time(T), fluent(in(C,P)), fluent(in(C,P')), P != P', holds(in(C, P'), T).  
holds(neg(in(C, P)), T)         :-  time(T), fluent(holding(K, C)), fluent(in(C, P)), holds(holding(K, C), T). 
holds(neg(in(C, P)), T)         :-  time(T), fluent(loaded(R, C)), fluent(in(C, P)), holds(loaded(R, C), T). 

%% action move a robot from a location to an adjacent non-occupied location  
action(move(R,L,L1))              :- robot(R),  location(L), location(L1), adjacent(L,L1).
executable(move(R,L,L1), T)  :- action(move(R,L,L1)), holds(at(R,L), T), holds(neg(occupied(L1)), T). 
holds(at(R, L1), T+1)               :- time(T), occ(move(R,L,L1), T).
holds(neg(occupied(L)), T+1)  :- time(T), occ(move(R,L,L1), T).

%% action: take a container with an empty crane from the top of a pile collocated with the crane 
action(take(C, K, P))               :-  container(C), crane(K), pile(P), location(L), attached(P, L), belong(K, L).
executable(take(C, K, P), T)   :-  time(T), action(take(C, K, P)), holds(empty(K), T), holds(top(C,P), T).
holds(holding(K, C), T+1)       :- time(T), occ(take(C, K, P), T). 
holds(top(C', P), T+1)             :-  time(T), occ(take(C, K, P), T), holds(on(C, C'), T). 
holds(neg(on(C, C')), T+1)      :-  time(T), occ(take(C, K, P), T), holds(on(C, C'), T). 

%% action: putdown a container held by a crane on the top of a pile collocated with the crane 
action(putdown(C, K, P))             :-  container(C), crane(K), pile(P), location(L), attached(P, L), belong(K, L).
executable(putdown(C, K, P), T) :-  time(T), action(putdown(C, K, P)), holds(holding(K, C), T).
holds(top(C, P), T+1)                   :-  time(T), occ(putdown(C, K, P), T). 
holds(empty(K), T+1)                   :-  time(T), occ(putdown(C, K, P), T). 
holds(neg(holding(K,C)), T+1)     :-  time(T), occ(putdown(C, K, P), T). 
holds(on(C, C'), T+1)                   :-  time(T), occ(putdown(C, K, P), T), holds(top(C', P), T). 

%% action: load a container held by a crane on an unloaded robot that is within the same location 
action(load(C, K, R))                    :-  container(C), crane(K), robot(R).
executable(load(C, K, R), T)        :-  time(T), action(load(C, K, R)), holds(unloaded(R), T), holds(at(R, L), T), holds(holding(K, C), T), belong(K, L).
holds(loaded(R, C), T+1)             :- time(T), occ(load(C, K, R), T). 
holds(empty(K), T+1)                   :- time(T), occ(load(C, K, R), T). 
holds(neg(holding(K,C)), T+1)     :-  time(T), occ(load(C, K, P), T). 
 
%% action: unload a container with an empty crane from a loaded robot that is within the same location 
action(unload(C, K, R))                :-  container(C), crane(K), robot(R).
executable(unload(C, K, R), T)    :-  time(T), action(unload(C, K, R)), holds(loaded(R, C), T), holds(at(R, L), T), holds(empty(K), T), belong(K, L).
holds(unloaded(R), T+1)              :- time(T), occ(unload(C, K, R), T). 
holds(holding(K, C), T+1)             :- time(T), occ(unload(C, K, R), T). 
holds(neg(loaded(R, C)), T+1)              :- time(T), occ(unload(C, K, R), T). 

%% defining the initial configuration  
holds(F, 0) :- initially(F). 
holds(neg(F), 0) :- fluent(F), not holds(F, 0).

%% inertia 
holds(F, T+1) :- time(T), fluent(F), holds(F, T), not holds(neg(F), T+1), T < len. 
holds(neg(F), T+1) :- time(T), fluent(F), holds(neg(F), T), not holds(F, T+1), T < len. 

%% action generation 
1{occ(A,T) : action(A)} 1 :- time(T), T < len. 
:- time(T), action(A), occ(A, T), not executable(A, T). 

\end{lstlisting}

\subsection{Generic Domain}

\begin{lstlisting}[language=clingo,caption= Generic Domain, label=lst:generic, 
mathescape=true,xleftmargin=.04\textwidth, breaklines=true]

%% working with numbers 
preconditions(0..m_pre). 
effects(1..m_effects). 
condition(0..m_c_effect). 

%% fluents 
fluent(1..fluents).

%% actions
action(1..actions).

%% defined literals 
lit(-L;L) :- fluent(L).
inconsistent(X) :- fluent(L), member(X,L), member(X,-L). 

%% generate randomly the number of executability condition for each action 
1 {precondition(A,pre(A),K) : preconditions(K)} 1 :- action(A). 

K {member(pre(A), L) : lit(L)} K :- action(A), precondition(A,pre(A),K), K>0.

%% the number of elements in the executability condition cannot be greater than the number of fluents 
:- action(A), precondition(A,pre(A),K), K > fluents.  

%% executability condition must be consistent 
:- action(A), lit(L), inconsistent(pre(A)). 

%% generate randomly the number of effects for each action 
1 {n_effects(A,K) : effects(K)} 1 :- action(A). 

%% for each effect, generates a consequence 
1 {effect(A, e(A,N), L) : lit(L)} 1 :- action(A), n_effects(A,K), effects(N), N <= K. 

%% determining the number of preconditions associated to the effect 
1 {effect(A, c(A,N), M) :  condition(M)} 1 :- action(A), n_effects(A,K), effects(N), N <= K. 

%% generating the preconditions
M {member(c(A,N), L) : lit(L)} M :- action(A), effects(N), effect(A, c(A,N), M), M > 0.

%% preconditions must be consistent 
:-  action(A), effects(N), effect(A, c(A,N), _), inconsistent(c(A,N)). 

%% checking consistency 
incompatible(X,Y) :- fluent(F), member(X, F), member(Y,-F).  
incompatible(X,Y) :- fluent(F), member(X, -F), member(Y,F).  
conflict(F,-F)          :- fluent(F). 
conflict(-F,F)          :- fluent(F). 
:-  action(A), effect(A, c(A,N), K), effect(A, c(A,N'), K'), N != N', 
    not incompatible(c(A,N), c(A,N')), 
    effect(A, e(A,N), L), effect(A, e(A,N'), L'), conflict(L,L').
:-  action(A),  
    effect(A, e(A,N), L), effect(A, e(A,N), L'), conflict(L,L').

% Counting the total number of actions with non-empty precondition 
nonempty_pre(T) :- T = #count {A : precondition(A, pre(A), K), K > 0}.

% making sure that the percentage is at least 1/3 -- change T*3 to something that you like 
% T*x means at least #actions/x must have non-empty precondition
:- nonempty_pre(T), actions > T*3. 

% This is to consider the total number of action effects 
 
nonempty_conditional_effects(T) :- T = #count {K,A :  action(A), effect(A, c(A, _), K), K > 0}. 
total_number_effects(T) :- T = #sum {K, A : action(A), n_effects(A,K)}. 

% making sure that the percentage is at least 1/3 -- change C*3 to something that you like 
% C*x means at least there must be at least #effects/x must have non-empty condition
:- nonempty_conditional_effects(C), total_number_effects(T), T > C*3. 
\end{lstlisting}


\end{document}